\documentclass[journal, 11pt, journal]{IEEEtran}
\onecolumn

\usepackage{graphicx,amsmath,amsfonts,amscd,amssymb,bm,url,color,latexsym}
\usepackage{subcaption}
\usepackage{bbm}
\usepackage{microtype}
\usepackage{algorithm}
\usepackage{algorithmicx}
\usepackage{algpseudocode}
\usepackage{hyperref}
\usepackage{verbatim}
\usepackage{framed}
\usepackage{comment}
\usepackage{tikz}
\usepackage{fge}
\usepackage[margin=0.9in]{geometry}
\usepackage{enumitem}
\usepackage[numbers,sort&compress]{natbib}
\usepackage{cleveref}
\usepackage{multirow}
\usepackage{tablefootnote}
\usepackage{comment}
\usepackage[affil-it]{authblk}

\usepackage{wrapfig}
\allowdisplaybreaks
\numberwithin{equation}{section}
\hypersetup{
    colorlinks=true,%
    citecolor=blue,%
    filecolor=blue,%
    linkcolor=blue,%
    urlcolor=blue
}

\newtheorem{theorem}{Theorem}[section]
\newtheorem{lemma}[theorem]{Lemma}

\def \endprf{\hfill {\vrule height6pt width6pt depth0pt}\medskip}

\newcommand{\mb}{\boldsymbol}
\newcommand{\mc}{\mathcal}

\newcommand{\bb}{\mathbb}

\newcommand{\eps}{\varepsilon}

\newcommand{\norm}[2]{\left\| #1 \right\|_{#2}}
\newcommand{\abs}[1]{\left| #1 \right|}
\newcommand{\innerprod}[2]{\left\langle #1,  #2 \right\rangle}

\definecolor{royalpurple}{rgb}{0.47, 0.32, 0.66}
\definecolor{lava}{rgb}{0.81, 0.06, 0.13}
\definecolor{darkblue}{rgb}{0, 0, 0.66}

\renewcommand{\eps}{\varepsilon}
\newcommand{\R}{\bb R}

\newcommand{\indicator}[1]{\mathbbm 1_{#1}}

\newcommand{\Brac}[1]{\left\lbrace #1 \right\rbrace}

\newcommand{\paren}[1]{ \left( #1 \right) }

\DeclareMathOperator{\dist}{dist}

\DeclareMathOperator{\sign}{sign}

\DeclareMathOperator{\grad}{grad}
\DeclareMathOperator{\Hess}{Hess}

\DeclareMathOperator*{\argmin}{argmin}

\makeatletter
\def\@IEEEsectpunct{\ \,}
\def\paragraph{\@startsection{paragraph}{4}{\z@}{1.5ex plus 1.5ex minus 0.5ex}%
{0ex}{\normalfont\normalsize\textbf}}
\makeatother

\newcommand{\ol}{\overline}

\newcommand{\edited}[1]{{\color{blue}{#1}}}

\newcommand{\zz}[1]{{\color{red}Zhihui: #1}}

\begin{document}

\title{Finding the Sparsest Vectors in a Subspace: Theory, Algorithms, and Applications}

\author[1,*]{Qing Qu\thanks{\normalsize * Both authors contribute equally to this work.}}
\author[2,*]{Zhihui Zhu}
\author[3]{Xiao Li}
\author[4]{Manolis C. Tsakiris}
\author[5]{John Wright}
\author[6]{Ren\'e Vidal}

\affil[1]{Center for Data Science, New York University}
\affil[2]{Department of Electrical and Computer Engineering, University of Denver}
\affil[3]{Department of Electronic Engineering, Chinese University of Hong Kong}
\affil[4]{School of Information Science and Technology, ShanghaiTech University}
\affil[5]{Department of Electrical Engineering \& Data Science Institute, Columbia University}
\affil[6]{Mathematical Institute for Data Science, Johns Hopkins University}


\maketitle

\begin{abstract}
The problem of finding the sparsest vector (direction) in a low dimensional subspace can be considered as a homogeneous variant of the sparse recovery problem, which finds applications in robust subspace recovery, dictionary learning, sparse blind deconvolution, and many other problems in signal processing and machine learning. However, in contrast to the classical sparse recovery problem, the most natural formulation for finding the sparsest vector in a subspace is usually nonconvex. In this paper, we overview recent advances on global nonconvex optimization theory for solving this problem, ranging from geometric analysis of its optimization landscapes, to efficient optimization algorithms for solving the associated nonconvex optimization problem, to applications in machine intelligence, representation learning, and imaging sciences. Finally, we conclude this review by pointing out several interesting open problems for future research.

\end{abstract}

\section{Introduction}\label{sec:intro}

Nonconvex optimization problems are \emph{ubiquitous} in signal processing and machine learning \cite{jain2017non,sun2019link}. However, for general nonconvex problems, even finding a local minimizer is NP-hard \cite{murty1987some}. 
While one may consider convex relaxations \cite{candes2008introduction,candes2009exact,candes2011robust,candes2015phase} and resort to the rich literature of convex optimization \cite{nesterov2013introductory,boyd2004convex}, such convex relaxations usually scale poorly with respect to the dimension of the data, and often provably fail for problems with nonlinear models. Nonetheless, recent advances in phase retrieval~\cite{netrapalli2013phase,sun2016geometric,candes2014wirtinger,chen2015solving}, phase synchronization~\cite{boumal2016nonconvex,liu2017estimation}, blind deconvolution \cite{lee2018fast,li2018rapid}, dictionary learning \cite{agarwal2013learning,sun2015complete},  matrix factorization \cite{burer2003nonlinear,burer2005local,jain2013low,tu2016low,sun2016guaranteed,bhojanapalli2016global,ge2016matrix,zhu2017GlobalOptimality,chi2018nonconvex}, tensor decomposition \cite{ge2015escaping}, etc., reveal that the optimization landscapes of nonconvex problems often have \emph{benign} geometric properties. 
The underlying benign geometric structure can be local or global, ensuring fast convergence of iterative algorithms to target solutions. Specifically:
\begin{figure*}[!htbp]
\centering
\begin{minipage}[c]{0.28\textwidth}
\centering
\includegraphics[height=1.5in]{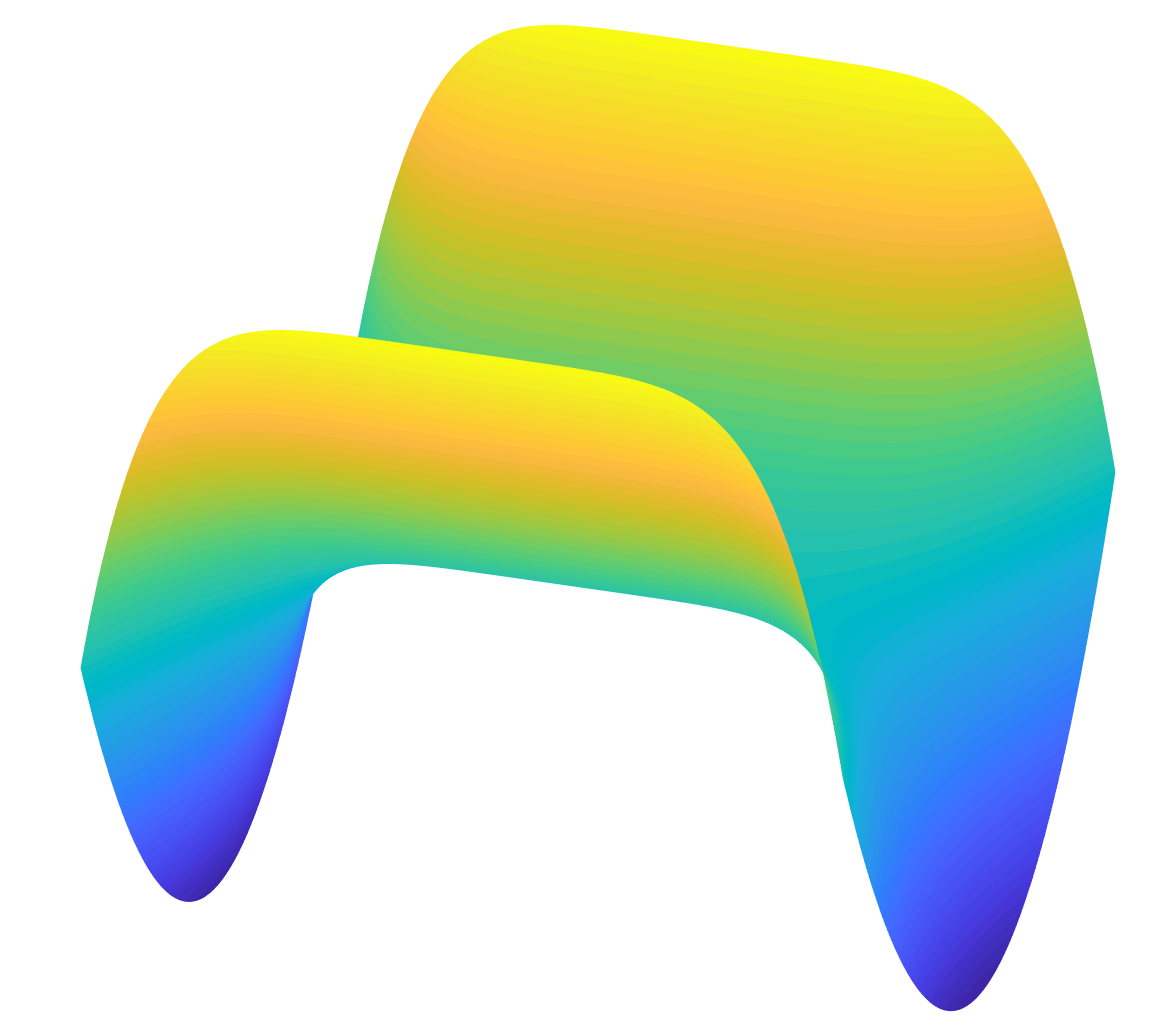}
\subcaption{Flat saddle points}\label{subfig:flat-saddles}
\end{minipage}
\hspace{0.5in}
\begin{minipage}[c]{0.28\textwidth}
\centering
\includegraphics[height=1.5in]{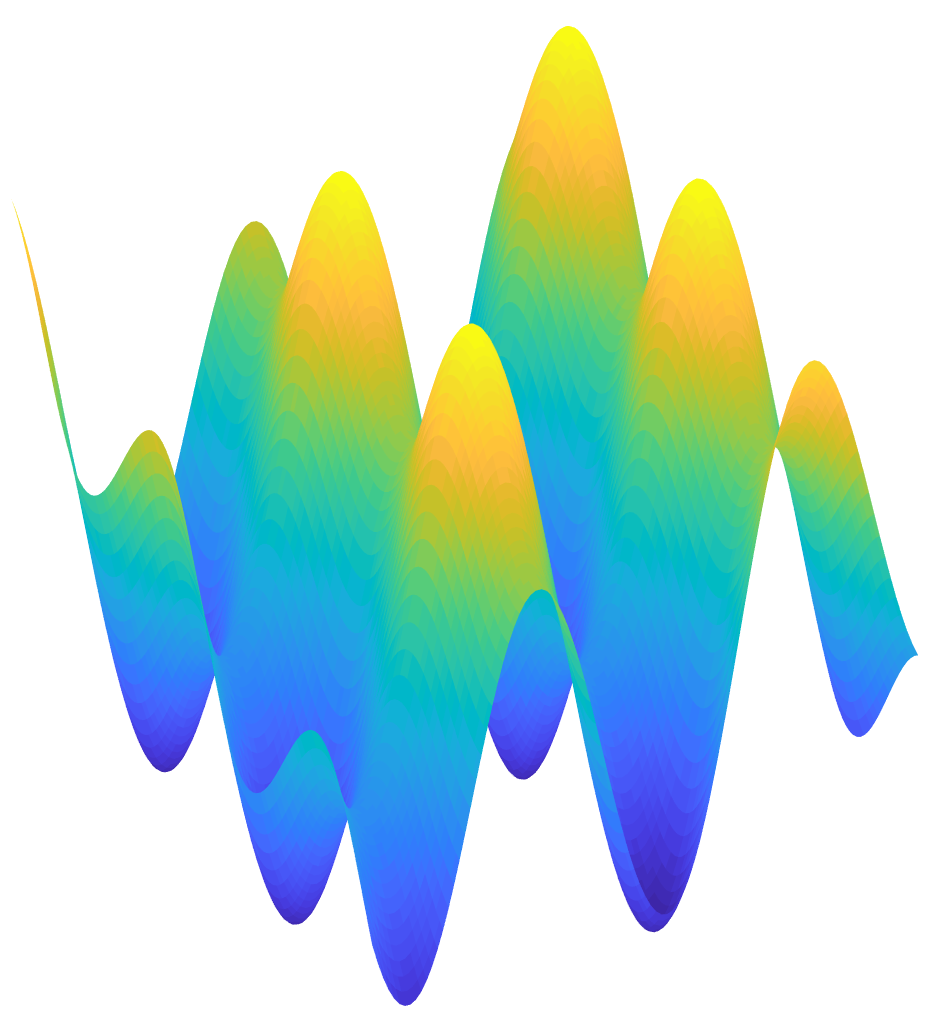}
\subcaption{Spurious local minimizers}\label{subfig:spurious-local}
\end{minipage}
\caption{\textbf{Two worst case scenarios in nonconvex optimization.}}
\label{fig:bad-landscape}
\end{figure*}
\begin{enumerate}[leftmargin=*]
\item \emph{Benign Local Geometry.} In many cases, there often exists a sufficiently large \emph{basin of attraction} around the target solutions, within which a local search algorithm converges rapidly to the solution.
\item \emph{Benign Global Geometry.} Problem specific symmetry structures induce a \emph{benign global optimization landscape} --- that there are \emph{no} flat saddle points or spurious local minima (see \Cref{fig:bad-landscape}) --- ensures global convergence of iterative algorithms from \emph{random} or \emph{arbitrary} initializations~\cite{sorensen1982newton,nesterov2006cubic,lee2016gradient,ge2015escaping,li2019alternating}.
\end{enumerate}

In this paper, we provide a comprehensive review of recent advances on nonconvex optimization methods for \emph{finding the sparsest vectors in a subspace} \cite{qu2016finding,barak2014rounding,demanet2014scaling,hopkins2016fast}. Namely, given data $\mb Y \in \bb R^{n \times p}$ whose rows form an $n$-dimensional subspace $\mc S \subseteq \R^p \;(n \ll p)$, can we \emph{efficiently} find the \emph{sparsest} nonzero vector in $\mc S$ (up to scalings)? Mathematically speaking, can we efficiently solve 
\begin{align}\label{eqn:problem-primal}
   \min_{\mb q\in\R^{n}}\; \norm{  \mb Y^\top \mb q}{0},\qquad \text{s.t.}\quad \mb q \not = \mb 0,
\end{align}
so that $\mb Y^\top \mb q$ is the sparsest vector\footnote{Here, $\mathrm{row}(\mb Y)$ denotes the row subspace of $\mb Y$, i.e., the subspace $\mathrm{row}(\mb Y)$ is spanned by row vectors of $\mb Y$.} in $\mc S = \mathrm{row}(\mb Y)$? Here, the nonzero constraint $\mb q \ne \mb 0$ avoids the \emph{trivial} sparse solution $\mb q=\mb 0$ which arises simply because a subspace $\mc S$ passes through the origin $\mb 0$. In the meanwhile, it should be noted that the problem can also be considered as a homogeneous variant of the sparse recovery problem \cite{donoho2006compressed,candes2008introduction}, in the sense that the problem \eqref{eqn:problem-primal} can be \emph{equivalently} formulated as  
\begin{align}\label{eqn:problem-dual}
\min_{\mb x} \; \norm{\mb x}{0}, \qquad \text{s.t.} \quad \mb A \mb x \;=\; \mb 0, \quad \mb x \;\ne\; \mb 0,
\end{align} 
where the rows of $\mb A \in \R^{\left(p - n\right) \times p}$ form a basis of the orthogonal complement of $\mc S$ so that \eqref{eqn:problem-dual} can be viewed as a \emph{dual} formulation of \eqref{eqn:problem-primal}. However, in contrast to the classical sparse recovery problem which finds the sparsest vector with $\mb A\mb x = \mb b$ and $\mb b \not=\mb 0$ \cite{donoho2006compressed,candes2008introduction,elad2010sparse,eldar2012compressed,foucart2013invitation}, solving \eqref{eqn:problem-dual} has caught less attention and has not been well-studied albeit its importance in many applications in signal processing and machine learning as we discuss in \Cref{sec:motivation}. One major reason is due to our limited understandings on the computational properties of solving the nonconvex problem \eqref{eqn:problem-dual}. Different from classical sparse recovery problems, where convex relaxations perform near optimally for broad classes of designs of $\mb A$~\cite{candes2005decoding, donoho2006most}, it has been known for decades that the basic problem \eqref{eqn:problem-dual} is NP-hard for an arbitrary subspace $\mc S$~\cite{mccormick1983combinatorial, coleman1986null}. Even if we relax the $\ell^0$-norm objective with a convex surrogate, the nonzero constraint $\mb x \not = \mb 0 $ still makes the problem \emph{inherently} nonconvex. It is only very recently that efficient computational surrogates with \emph{nontrivial} recovery guarantees have been discovered and studied for specific instances \cite{spielman2013exact,qu2016finding,sun2016complete-i,wang2016blind,rahmani2017innovation,li2018global,tsakiris2018dual,zhu2018dual}. In this paper, we survey several important aspects of recent advances on nonconvex optimization methods for solving the problem of finding the sparsest vector in a subspace, ranging from landscape analysis, to efficient optimization methods, to applications.

\vspace{0.05in}

\noindent \textbf{Paper organization.} The rest of the paper is organized as follows. In \Cref{sec:motivation}, we show that several fundamental problems in signal processing and machine learning can be reduced to the task of finding the sparsest vector in a subspace. In \Cref{sec:problem}, we introduce natural nonconvex relaxations of \eqref{eqn:problem-primal} with computational guarantees. In \Cref{sec:geometry} we provide a systematic overview of the geometric analysis on nonconvex optimization landscapes, based on which nonconvex algorithms have recently led to efficient solutions and new performance guarantees that we discuss in \Cref{sec:optimization}. We demonstrate the broad applications in \Cref{sec:application}. Finally, we close this review by discussing several open problems in \Cref{sec:conclusion}.

\section{Motivations}\label{sec:motivation}

Variants of the task of finding the sparsest vector in a subspace take several forms in many applications of modern signal processing and machine learning. In this section, we survey several fundamental problems to demonstrate its importance, where all the problems can be reduced to solving \eqref{eqn:problem-primal}, with different structures of the subspace $\mc S = \mathrm{row}(\mb Y)$. 

\paragraph{Robust subspace recovery \cite{jolliffe2011principal,candes2011robust,lerman2011robust,xu2012robust,lerman2018overview}.} 

Fitting a linear subspace to dataset corrupted by outliers is a fundamental problem in machine learning and statistics, primarily known as robust principal component analysis (PCA) \cite{vidal2016generalized}, or robust subspace recovery (RSR) \cite{lerman2018overview}. Given the dataset $\mb Y$ of the form
\begin{align}
  \underset{  \textbf{\color{lava} data} }{\mb Y} \quad =\quad \Big[\; \underset{ \textbf{\color{lava} inliers} }{ \mb X } \, \, \underset{\textbf{\color{lava} outliers} }{ \mb O } \;  \Big] \; \underset{ \textbf{\color{lava} permutation} }{\mb \Gamma} \;\in \;\bb R^{n\times p},
\label{eq:DPCP Y}\end{align}
where the columns of $\mb X \in \bb R^{n \times p_1}$ form inlier points spanning a subspace $\mc S_{\mb X}$, the columns of $\mb O \in \bb R^{n \times p_2}$ are outlier points with no linear structure, and $\mb \Gamma$ is an unknown permutation, the goal here is to recover the inlier subspace $\mc S_{\mb X}$, or equivalently to cluster the points into inliers and outliers. It is well-known that the presence of outliers can severely affect the quality of the solutions obtained by the classical PCA approach \cite{vidal2016generalized}. This challenge can be conquered by finding the sparsest vector in
$\mc S= \mathrm{row}(\mb Y)$ via solving \eqref{eqn:problem-primal}, which returns a normal vector\footnote{A normal vector of a subspace is a nonzero vector that is orthogonal to all points in the subspace.} of the subspace $\mc S_{\mb X}$ \cite{tsakiris2018dual}, producing a hyperplane containing all columns of $\mb X$. This approach is called \emph{dual principal component pursuit} (DPCP) \cite{tsakiris2018dual,zhu2018dual,pmlr-v97-ding19b}, which can be viewed as a \emph{dual} method of classical ways of solving robust subspace recovery problems \cite{lerman2018overview}. The DPCP has led to new recovery guarantees, which can deal with more number of outliers than traditional methods \cite{tsakiris2018dual,zhu2018dual,pmlr-v97-ding19b}. Moreover, it also shows the potential for tackling multiple subspace fitting~\cite{tsakiris2017hyperplane}.

\paragraph{Learning sparsely-used dictionaries \cite{olshausen1996emergence,aharon2006k,elad2010sparse,rubinstein2010dictionaries}.} 
Dictionary learning (DL) aims to learn the underlying compact representation from the data $\mb Y$, which finds many applications in signal/imaging processing, machine learning, and computer vision \cite{aharon2006k,elad2006image,rubinstein2010dictionaries,wright2010sparse,mairal2014sparse}. Mathematically speaking, the problem is to factorize the data 
\begin{align}
  \underset{  \textbf{\color{lava} data} }{\mb Y} \quad = \quad  	 \underset{  \textbf{\color{lava} dictionary} }{\mb A} \;\; \underset{  \textbf{\color{lava} sparse code} }{\mb X}
\label{eq:DL Y}\end{align}
into a compact representation dictionary $\mb A$ and sparse coefficient matrix $\mb X$. Such representations naturally allow signal compression \cite{aharon2006k}, and also facilitate efficient signal acquisition \cite{bruckstein2009sparse}, denoising \cite{elad2006image}, and classification \cite{mairal2011task} (see relevant discussion in~\cite{mairal2014sparse}). In particular, when the dictionary $\mb A$ is \emph{complete}\footnote{Complete means that the dictionary $\mb A$ is square and invertible. For a proper conditioned dictionary, the complete DL can be approximately reduced to orthogonal DL via preconditioning or whitening of the data \cite{sun2016complete-i}.}, the authors in \cite{spielman2013exact,qu2016finding,sun2016complete-i} showed that the DL problem can be reduced to finding the sparsest vector in the subspace $\mc S=\mathrm{row}(\mb Y)$: by solving \eqref{eqn:problem-primal}, the solution $\mb Y^\top \mb q$ is expected to be one row of the sparse matrix $\mb X$. Based on this, the full matrices $\paren{\mb A,\mb X}$ can be recovered via extra techniques such as deflation \cite{spielman2013exact}. For complete DL, this approach has led to new theoretical and algorithmic advances~\cite{spielman2013exact,sun2016complete-i,sun2016complete-ii,gilboa2019efficient,bai2018subgradient}.

\paragraph{Sparse blind deconvolution \cite{zhang2017global,kuo2019geometry,campisi2016blind,li2018bilinear,shi2019manifold}.} Sparse blind deconvolution is a classical inverse problem that ubiquitously appears in various areas of digital communication \cite{moulines1995subspace}, signal/image processing \cite{kundur1996blind,levin2009understanding}, neuroscience \cite{vogelstein2010fast,pnevmatikakis2016simultaneous}, geophysics \cite{kazemi2014sparse}, and more. Given multiple measurements $\Brac{\mb y_i}_{i=1}^p$ in the form of the circulant convolution\footnote{Here, we use $\circledast$ to denote circulant convolution, which can be efficiently implemented via fast Fourier transform \cite{li2018global,lau2019short}. It should also be noted that any linear convolution can be rewrite a circulant convolution via properly zero padding the vectors.} 
\begin{align}\label{eqn:mcs-bd-problem}
    \underset{  \textbf{\color{lava} measurements} }{\mb y_i} \quad = \quad \underset{ \textbf{\color{lava} kernel} }{\mb a_0} \; \circledast \; \underset{ \textbf{\color{lava} sparse signal} }{\mb x_i},\qquad 1 \;\leq\; i \;\leq\; m,
\end{align}
the \emph{multichannel sparse blind deconvolution} (MCS-BD) problem \cite{wang2016blind,li2018global,qu2019blind,shi2019manifold} aims to simultaneously recover the unknown kernel $\mb a_0$ and sparse signals $\Brac{\mb x_i}_{i=1}^p$. Notice that the circulant convolution \eqref{eqn:mcs-bd-problem} can be rewritten in the matrix-vector form with\footnote{Here, any vector $\mb v\in \bb R^n$, we use $\mb C_{v} \in \bb R^{n \times n}$ to denote corresponding circulant matrix generated from $\mb v$.} $\mb C_{\mb y_i} = \mb C_{\mb a} \mb C_{\mb x_i} $. Thus, by concatenating all the measurements, we can write the problem in a similar form of complete DL in the sense that
\begin{align*}
   \underbrace{\begin{bmatrix}
   	\mb C_{\mb y_1} & \cdots & \mb C_{\mb y_p}
   \end{bmatrix} }_{ \mb Y } \quad=\quad   \mb C_{\mb a_0}  \quad \underbrace{\begin{bmatrix}
   	\mb C_{\mb x_1} & \cdots & \mb C_{\mb x_p}
   \end{bmatrix} }_{ \mb X }.
\end{align*}
When the kernel $\mb a_0$ is invertible\footnote{In other words, we assume that its circulant matrix $\mb C_{\mb a_0}$ is invertible.}, per our discussion for complete DL, this implies that we can solve the MCS-BD problem by finding the sparsest vector in $\mc S = \mathrm{row}(\mb Y)$ in a similar fashion. This discovery has recently led to new guaranteed, efficient methods for solving MCS-BD under general settings \cite{li2018global,qu2019blind,shi2019manifold}.

\vspace{-0.05in}

\paragraph{Other problems.} Variants and generalizations of finding the sparsest vectors in a subspace problem also appear in orthogonal $\ell^1$ regression~\cite{spath1987orthogonal}, sparse PCA \cite{zou2006sparse,Aspremont07sparse}, numerical linear algebra \cite{coleman1986null,gilbert86computing,gottlieb2010matrix}, applications regarding control and optimization~\cite{zhao2013rank}, nonrigid structure from motion~\cite{dai2012simple}, spectral estimation and the Prony's method~\cite{beylkin2005approximation}, blind source separation~\cite{zibulevsky2001blind}, graphical model learning~\cite{anandkumar2013overcomplete}, and sparse coding on manifolds~\cite{ho2013nonlinear}. Nonetheless, we believe the potential of seeking sparse/structured elements in a subspace is still largely unexplored, in spite of the cases we discussed here. We hope this review will bring more attention to this problem and inspire further application ideas of recent theoretical and algorithmic advances.


\section{Problem Formulation}\label{sec:problem}

 \begin{table*}  
\setlength{\tabcolsep}{11pt}
\renewcommand{\arraystretch}{1}
 \begin{center}
\resizebox{\textwidth}{!}{
 \begin{tabular}{c||c|c|c}
 \hline
 Name & Objective $\varphi(\cdot)$ &  (Sub)gradient $\nabla \varphi(\cdot)$ & Smoothness  \\ 
 \hline
 $\ell^1$-norm \cite{bai2018subgradient,zhu2018dual} &  $ \sum_k \abs{z_k} $ & $\sign\paren{\mb z} $ & nonsmooth \\ 
 \hline 
 Huber loss \cite{qu2019blind} & $  \sum_k  \left(\frac{z_k^2}{2\mu} + \frac{\mu}{2} \right) \indicator{  \abs{\mb z_k} < \mu } +  |z_k| \indicator{\abs{\mb z_k} \geq \mu }$ & $  \mb z /\mu  \indicator{  \abs{\mb z} < \mu }  + \sign(\mb z) \indicator{\abs{\mb z} \geq \mu } $ & $\mc C^1$-smooth \\
\hline
 pseudo-Huber \cite{kuo2019geometry} &  $  \mu \sum_k \sqrt{ 1 + (z_k/\mu)^2  } $ & $
 \mb z / \sqrt{ \mb z + \mu^2 } $ & $\mc C^\infty$-smooth \\
\hline
 Logcosh \cite{sun2016complete-i,shi2019manifold} &  $ \sum_k \mu\log \cosh(z_k/\mu)$ & $\tanh \paren{\mb z/\mu } $ & $\mc C^\infty$-smooth \\
\hline

\end{tabular}
}
 \caption{\textbf{Summary of Convex Surrogates $\varphi(\cdot)$ for $\ell^0$-norm.}}\label{tab:comparison-loss}
 \end{center}
\end{table*}



Per our discussion in \Cref{sec:intro}, to find the sparsest vectors in $\mc S = \mathrm{row}(\mb Y)$ the vanilla formulation \eqref{eqn:problem-primal} (or equivalently \eqref{eqn:problem-dual}) is NP-hard to solve. Therefore, we need to resort to certain relaxations of the problem such that the new problem is substantially easier to optimize and its global solutions are still \emph{close} to the expected target solutions. Similar to the idea of solving the sparse recovery problem \cite{candes2005decoding,candes2008introduction}, one natural idea is to replace $\ell^0$-norm with any sparsity promoting convex surrogate $\varphi(\cdot)$ (see \Cref{tab:comparison-loss} for an illustration, we will discuss the choices in \Cref{sec:optimization}). However, the nonconvex constraint $\mb q \not = \mb 0$ still makes the problem inherently difficult to optimize. Nonetheless, since we only hope to find the sparsest vector up to a scaling, it is natural to consider replacing $\mb q \not = \mb 0$ by certain unit norm constraints on $\mb q$.

\vspace{0.05in}

\noindent \textbf{Limitation of convex relaxations.} In the context of the dictionary learning problem, the work \cite{spielman2013exact} first considered an $\ell^1$-minimization problem constrained with $\norm{\mb q}{ \infty } =1 $, introducing a convex relaxation of \eqref{eqn:problem-primal} with a sequence of linear programs:
\begin{align}\label{eqn:l-infty-relaxation}
\ell^1/\ell^\infty \text{ Relaxation:}\qquad  \min_{\mb q} \; \varphi \paren{ \mb Y^\top \mb q },\quad \text{s.t.} \quad q(i) = 1,
\end{align}
for some $i\in\{1,\ldots,n\}$. Here $\varphi(\cdot) = \norm{\cdot}{1} $ as shown in \Cref{tab:comparison-loss}. The solutions of \eqref{eqn:l-infty-relaxation} are exactly the target sparse vectors up to a scaling when the subspace $\mc S$ is spanned by a set of random sparse basis vectors. However, this result provably breaks down merely when the sparsity level of each base vector is beyond\footnote{Here, $\theta$ denotes the probability of one entry being nonzero.} $\theta \in \mc O\paren{ 1/\sqrt{n}}$, while convex relaxation for standard sparse approximation problems can handle much higher sparsity levels $\theta = \Omega(1)$ \cite{candes2005decoding, donoho2006most}. For the problem \eqref{eqn:l-infty-relaxation}, the same sparsity threshold is also observed for a simpler \emph{planted sparse vector} (PSV) model, where there is a single sparse vector embedded in an otherwise random subspace $\mc S$ \cite{demanet2014scaling}. Moreover, for both models the most natural semidefinite programming (SDP) relaxation \cite{qu2016finding} also breaks down at exactly the same threshold\footnote{This breakdown behavior is again in sharp contrast to the standard sparse approximation problem (with $\mb b \ne \mb 0$), in which it is possible to handle very large fractions of nonzeros (say, $\theta = \Omega(1/\log n)$, or even $\theta = \Omega(1)$) using a very simple $\ell^1$ relaxation~\cite{candes2005decoding, donoho2006most}}. Unfortunately, numerical simulations confirm that these results are essentially sharp, so that one might naturally ask: {\em is $\theta \in \mc O\paren{1/\sqrt{n}}$ the best we can do with efficient, guaranteed algorithms?}

Remarkably, this is not the case. Recently, a new rounding technique for sum-of-squares (SoS) relaxations indicates that the sparsest vector can be recovered even when $\theta = \Omega(1)$  \cite{barak2015dictionary}.  Unfortunately, the runtime of this approach is a high-degree polynomial in data dimension so that the result is mostly of theoretical interest. Therefore, the question remains legitimate: \emph{Is there a practical algorithm that provably recovers a sparse vector with $\theta = \Omega(1)$ portion of nonzeros from a generic subspace $\mc S$?}

\vspace{0.05in}
 
\noindent \textbf{Efficient solutions via nonconvex optimization.}  This challenge has been addressed by recent advances on nonconvex optimization, where the work in \cite{qu2016finding} first considered a \emph{nonconvex} relaxation of \eqref{eqn:problem-primal},
\begin{align} \label{eqn:L1-ncvx}
\ell^1/\ell^2 \text{ Relaxation:}\qquad \min_{\mb q} \; f(\mb q) \;:=\; \varphi \paren{ \mb Y^\top \mb q }, \qquad \text{s.t.} \quad \norm{\mb q}{2} \;=\; 1,
\end{align}
which relaxes $\mb q \ne \mb 0$ by a \emph{nonconvex} spherical constraint $\mb q \in \bb S^{n-1}$. Intuitively, the sphere $\bb S^{n-1}$ is a homogeneous manifold so that it could potentially handle higher sparsity levels. Indeed, for an idealized PSV model, the result in \cite{qu2016finding} showed that there is a simple, efficient nonconvex optimization method that provably recovers the sparsest vector even with $\theta = \Omega(1)$, breaking the $\theta \in \mc O\paren{ 1/\sqrt{n}}$ sparsity barrier. Subsequent work \cite{sun2016complete-i,sun2016complete-ii} showed that the same sparsity level can also be achieved with efficient methods for complete DL. Inspired by these results, optimizing variants of the nonconvex formulation \eqref{eqn:L1-ncvx} has led to new performance guarantees in robust subspace recovery \cite{tsakiris2018dual,zhu2018dual} and sparse blind deconvolution \cite{li2018global,qu2019blind,shi2019manifold}. Nonetheless, as the problem formulation in \eqref{eqn:L1-ncvx} is nonconvex, it naturally raises the following question: {\em what are the underlying principles for efficiently solving these nonconvex problems to target solutions?}

\section{Geometry and Optimization Landscapes}\label{sec:geometry}

\begin{figure*}[t]
\centering
\begin{minipage}[c]{0.22\textwidth}
\centering
\includegraphics[width = \linewidth]{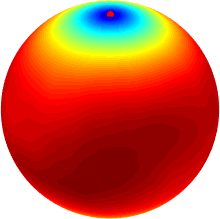}
\subcaption{PSV}\label{subfig:landscape-psv}
\end{minipage}
\hspace{0.01\textwidth}
\begin{minipage}[c]{0.22\textwidth}
\centering
\includegraphics[width = \linewidth]{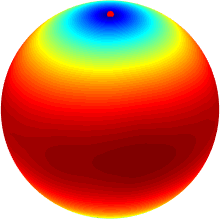}
\subcaption{DPCP}\label{subfig:landscape-dpcp}
\end{minipage}
\hspace{0.01\textwidth}
\begin{minipage}[c]{0.22\textwidth}
\centering
\includegraphics[width = \linewidth]{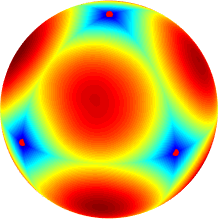}
\subcaption{ODL}\label{subfig:landscape-dl}
\end{minipage}
\hspace{0.01\textwidth}
\begin{minipage}[c]{0.22\textwidth}
\centering
\includegraphics[width = \linewidth]{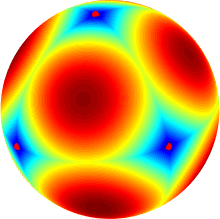}
\subcaption{MCS-BD}\label{subfig:landscape-mcsbd}
\end{minipage}
\caption{\textbf{Plots of optimization landscapes of \eqref{eqn:L1-ncvx} over the sphere for different problems in 3D.} From the left to right: (a) planted sparse vector \cite{qu2016finding}, (b) robust subspace recovery \cite{tsakiris2018dual}, (c) orthognal dictionary learning \cite{sun2016complete-i}, and (d) sparse blind deconvolution \cite{li2018global,qu2019blind}. The colder color value means smaller function value, and vice versa. The red dots correspond to the target solutions.  }
\label{fig:opt-landscape}
\end{figure*}

In the following, we demystify the recent success of nonconvex approaches, by reviewing recent advances on geometric studies of the nonconvex optimization landscapes, towards providing a unified view for solving a broader class of nonconvex optimization problems. Correspondingly, in the next section (\Cref{sec:optimization}) we show how to exploit these benign geometric properties plus extra ingredients to develop efficient nonconvex optimization methods, efficiently solving \eqref{eqn:L1-ncvx} to target (global) solutions.

\subsection{Some Basic Facts}

\vspace{0.05in}

\noindent \textbf{Basic notations.} First, we introduce some basic notations for studying the global optimization properties. Let $\mc Q_\star \subset \bb S^{n-1} $ be the set of target solutions. To measure the distance between a vector $\mb q \in \bb S^{n-1}$ and the set $\mc Q_\star$, we introduce the following metric defined in the Eucildean space
\begin{align*}
    \dist(\mb q,{\mc Q}_\star) \;:=\; \inf_{\mb a\in {\mc Q}_\star} \norm{ \mb q - \mb a}{2}.
\end{align*}
Accordingly, we define the set
\begin{align*}
    \mc B(\mb q,\mc Q_\star) \;:=\; \Brac{ \mb q \in \bb S^{n-1} \;\mid\; \dist\paren{ \mb q, \mc Q_\star } \leq  \eps  } 
\end{align*}
that contains all the points on the sphere that are $\eps$-close to $\mc Q_\star$.

\vspace{0.05in}

\noindent \textbf{Riemannian derivatives.}
Since we are solving a nonconvex optimization problem \eqref{eqn:L1-ncvx} that is constrained over a Riemannian manifold $\bb S^{n-1}$, to study the geometric properties of optimization landscape, we need formal definitions of the \emph{slope} (gradient) and \emph{curvature} (Hessian) of $f(\cdot)$ over the manifold. The sphere $\bb S^{n-1}$ is a smooth manifold embedded in $\bb R^n$; its {\em tangent space} $\mb q^\perp$ at the point $\mb q \in \bb S^{n-1} $ can be identified with 
\begin{align*}
    \mathrm{T}_{\mb q}\bb S^{n-1} \;=\; \Brac{ \; \mb v \in \bb R^n \; \mid\; \mb q^\top \mb v \;=\; 0  }.
\end{align*}
Thus, the projection onto the tangent space is given by $\mc P_{\mb q^\perp} = \mb I - \mb q \mb q^\top $. If $f$ is smooth, the slope of $f(\cdot)$ over the sphere (formally, the Riemannian gradient) is defined in the tangent space $\mathrm{T}_{\mb q}\bb S^{n-1}$, which is simply the component of the standard (Euclidean) gradient $\nabla f(\mb q)$ that is tangent to the sphere:
\begin{align*}
    \grad[ f ](\mb q) \;=\; \mc P_{\mb q^\perp} \nabla f(\mb q).
\end{align*}
When $f$ is nonsmooth, we can similarly introduce the corresponding Riemannian subgradient\footnote{We refer to \cite{yang2014optimality} for a formal defintion of Riemannian subgradient.
For a general nonsmooth function, the projection (onto the tangent space) of a subgradiment may not be a Riemannian subgradient. Fortunately, for problem that is regular (such as $f(\mb q) = \norm{\bm Y^\top \bm q}{1}$), according to \cite{yang2014optimality}, the Riemannian subgradient can be simply introduced by the projection of a subgradient.} 
\begin{align*}
    \partial_{\mathrm{R}} f(\mb q) \;=\; \mc P_{\mb q^\perp} \partial f(\mb q),
\end{align*}
where $\partial f(\mb q)$ is a particular choice of the subgradient of $f$. For instance, if  $f(\mb q) = \norm{\bm Y^\top \bm q}{1}$, we often choose $\partial f(\mb q) =  \bm Y\sign(\bm Y^\top \bm q)$ where $\sign(\cdot)$ is an element-wise sign operator that sets output to $0$ if the input is $0$. 

On the other hand, if\footnote{A function $f$ is said to be of $C^k$ if its k-th order derivative exists and is continuous.} $f \in \mc C^2$, the curvature of $f(\cdot)$ over the sphere is slightly more involved. For any direction $\mb \delta \in \mathrm{T}_{\mb q}\bb S^{n-1}$, the second derivative of $f(\cdot)$ at point $\mb q \in \bb S^{n-1}$ along the geodesic curve\footnote{A geodesic curve is the shortest path connecting two points on the manifold, which can be parameterized by an exponential map $\exp_{\mb q}\paren{t\delta}$. We refer readers to \cite{absil2009} for more technical details.} is given by $\mb \delta^\top \Hess[f](\mb q) \mb \delta$, where $\Hess[f](\mb q)$ is the \emph{Riemannian Hessian}  
\begin{align*}
    \Hess[f](\mb q) \;=\; \mc P_{\mb q^\perp} \Big( \; \underset{  \textbf{\color{lava} curvature of }f(\cdot) }{\nabla^2 f(\mb q) } -  \underset{  \textbf{\color{lava} curvature of the manifold} }{ \innerprod{ \mb q }{ \nabla f(\mb q) } \mb I } \; \Big) \mc P_{\mb q^\perp}.
\end{align*}
This expression contains two terms: (i) the first is the standard (Euclidean) hessian $\nabla^2 f$, which accounts for the curvature of the objective function $f$; (ii) the second term accounts for the curvature of the sphere itself. Thus, analogous to the case in Euclidean space, critical points can be characterized by $\grad[ f ](\mb q) = \mb 0 $ or $\mb 0 \in \partial_{\mathrm{R}} f(\mb q)$; curvatures can be studied through the eigenvalues of $\Hess [ f](\mb q) $.

\subsection{Local Geometry: Basins of Attraction Around Target Solutions}\label{subsec:local-geometry}

At the early stage of studying nonconvex optimization, people tend to believe that nonconvex problems \emph{only} have benign local geometric structures such that smart initializations are  needed. To guarantee a local search algorithm find a minimzer of a nonconvex problem, one natural idea is to show that there exist \emph{local basins of attraction} around the target solutions, in the sense that the function either has local \emph{strong convexity} or it satisfies certain \emph{regularity condition} around the target solutions. Therefore, to have guaranteed global optimization, people developed data-driven initialization by using spectral methods to initialize into the local basin such that descent methods efficiently converge to the target solutions. The initialization plus local algorithmic analysis has led to global guarantees for several important problems in signal processing and machine learning, such as generalized phase retrieval \cite{netrapalli2013phase,candes2015retrieval}, low rank matrix recovery \cite{chi2018nonconvex,li2018nonconvex}, tensor decomposition \cite{anandkumar2014tensor}, and blind deconvolution with subspace model \cite{li2019rapid}, and more.

In the context of finding the sparsest vector in a subspace, for a function $\varphi$ that is $\mc C^\infty$-smooth, the corresponding problem is \emph{locally} strongly convex in the sense that \cite{sun2016complete-i,li2018global}
\begin{align}\label{eqn:local-strong-convex}
    \Hess[f](\mb q) \;\; \succeq \;\; \alpha \cdot \mb P_{\mb q^\perp}, \quad  \forall \; \mb q\in {\cal B}\paren{{\cal Q}_\star,\epsilon_1},
\end{align}
where $\alpha>0$ is some scalar depending on the dimension of the problem.
However, for many nonconvex problems the regions that satisfy strong convexity are usually quite small (i.e., $\epsilon_1$ is small), that they only cover a small measure of the sphere. For problems such as complete DL \cite{sun2016complete-i} and sparse blind deconvolution \cite{li2018global}, it is often very difficult to initialize into the region ${\cal B}\paren{{\cal Q}_\star,\epsilon_1}$.
Moreover, the strong convexity condition also needs the function to be at least $\mc C^2$-smooth, which is quite stringent.  

A more general local condition is the so-called \emph{regularity condition}, which often ensures local convergences of descent methods within a region of much larger radius. For instance, for a consideration of nonsmooth $f$, the following regularity condition\footnote{The consideration of nonsmooth objective is only for the ease to resort to existing results \cite{bai2018subgradient,zhu2019linearly}, and the simplicity of presenting the regularity condition. } has been repeatedly discovered for many problems \cite{bai2018subgradient,zhu2019linearly}
\begin{align}
\innerprod{\mb q - {\cal P}_{{\cal Q}_\star}(\mb q)}{\partial_{\mathrm{R}} f(\mb q)} \;\geq\; \alpha \cdot \dist(\mb q,{\cal Q}_\star), \ \quad \ \forall \ \mb q\in {\cal B}({\cal Q}_\star,\epsilon_2).
\label{eq:RC-nonsmooth-0}
\end{align}

\begin{wrapfigure}{R}{0.37\textwidth}
\centering	\includegraphics[height=1.5in]{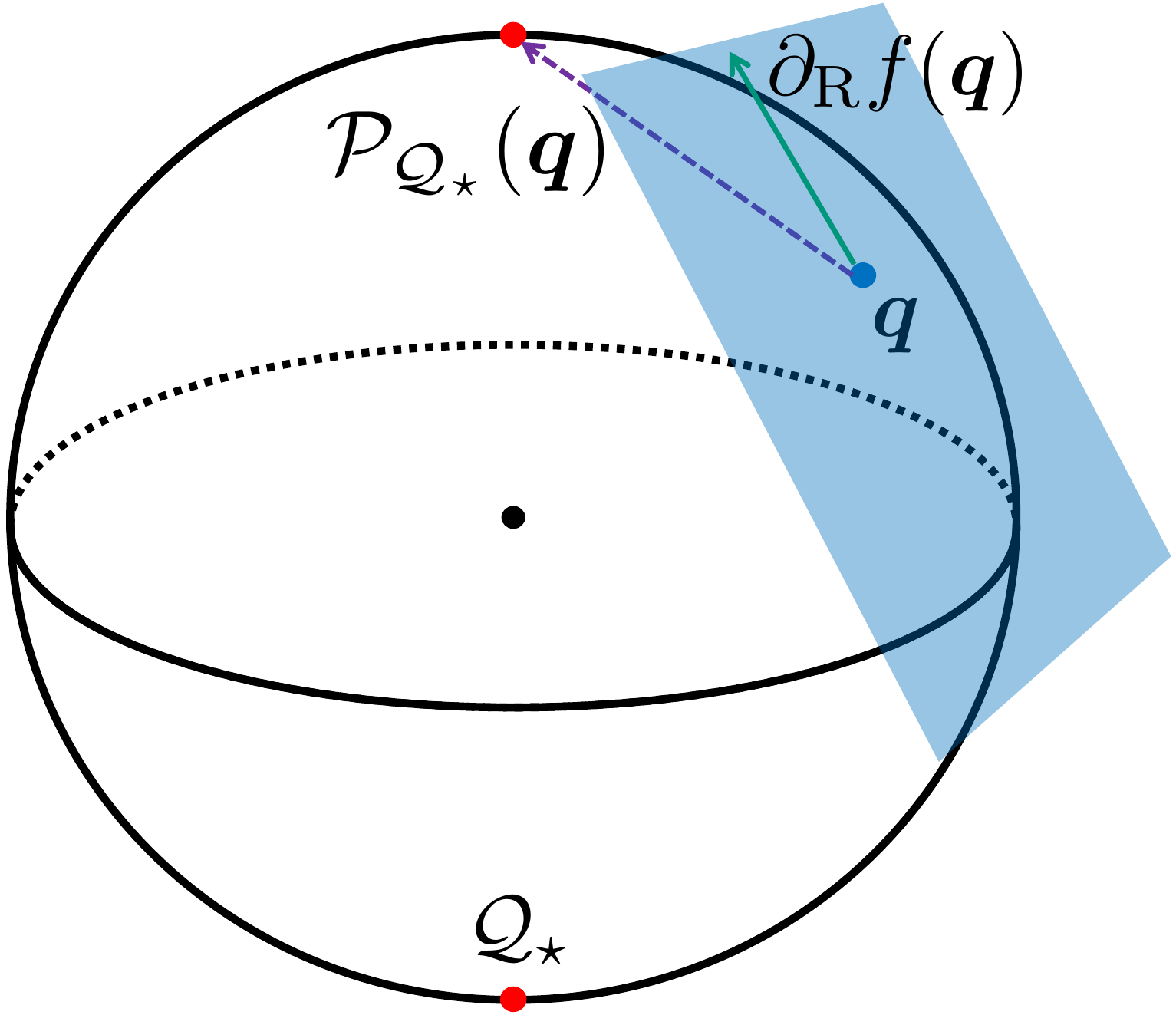}
	\caption{\textbf{Illustration of \Cref{eq:RC-nonsmooth-0}.} Red nodes denote ${\cal Q}_\star$, with the top one closest to ${\cal Q}_\star$. Inequality~\eqref{eq:RC-nonsmooth-0} requires the angle between ${\cal P}_{\cal Q^\star}(\mb q) -\mb q$ (purple arrow) and $-\partial_R f(\mb q)$ (blue arrow) to be sufficiently small.
	}\label{fig:describe RC}
\vspace{-.1in}
\end{wrapfigure}

As illustrated in \Cref{fig:describe RC}, the condition \eqref{eq:RC-nonsmooth-0} shows that the negative direction of the chosen Riemannian subgradient $\partial_{\mathrm{R}} f(\mb q)$ is aligned with the direction $\mb q - {\cal P}_{{\cal Q}_\star}(\mb q)$ pointing to the target solutions. In other words, this regularity condition will force the trajectory of (sub)gradient iterates getting closer to the target solutions when the step size is chosen appropriately, which we will discuss in more details in \Cref{sec:optimization}. Moreover, \eqref{eq:RC-nonsmooth-0} indicates a lower bound\footnote{This lower bound can be obtained by applying the Cauchy-Schwartz inequality to the left hand side of \eqref{eq:RC-nonsmooth-0}.} for the Riemannian subgradient $\norm{\partial_R f(\bm q)}{2}\ge \alpha$ for all $q\in {\cal B}({\cal Q}_\star,\epsilon_2)\setminus {\cal Q}_\star$. Thus, if \eqref{eq:RC-nonsmooth-0} holds for all the Riemannian subgradients of any $q\in {\cal B}({\cal Q}_\star,\epsilon_2)$, so that one can conclude there is no critical point other than the target solutions ${\cal Q}_\star$ in ${\cal B}({\cal Q}_\star,\epsilon_2)$. This property further implies the possibility of finding a target solution by not only the Riemannian subgradient method but also many other iterative algorithms (which will be described in \Cref{sec:optimization}) as long as they are initialized properly and can exploit this geometric property.

\vspace{0.1in}
\noindent \textbf{Minimal Example I: Robust Subspace Recovery.}  In the context of finding the sparsest vector in a subspace, we use the robust subspace recovery problem as an example to elaborate on the regularity condition \eqref{eq:RC-nonsmooth-0}. As illustrated in \Cref{sec:motivation}, given the dataset $\bm Y$ corrupted by outliers as $\mb Y = \begin{bmatrix} \bm X & \bm O\end{bmatrix} \bm \Gamma$, where $\mb X \in \bb R^{n \times p_1}$ are inliers generated from a subspace $\mc S_{\mb X}$,  $\mb O \in \bb R^{n \times p_2}$ are outliers with no linear structure, and $\bm \Gamma$ is an unknown permutation matrix, the goal is to recover the underlying inlier subspace $\mc S_{\mb X}$. Noting that estimating $S_{\mb X}$ is  equivalent to finding its orthogonal complement $\mc S_{\mb X}^\perp$, the DPCP approach \cite{manolis2015dualPCA} attempts to find one basis vector $\bm q \in \mc S_{\mb X}^\perp$ in each time. Once one basis vector is founded, we can then find another basis vector by removing the contribution from the previous one and repeat this process until finding all the basis vectors for $\mc S_{\mb X}^\perp$ (this is also called the deflation method).

Recall that if $\bm q$ is in the orthogonal complement subspace $\mc S_{\mb X}^\perp$, then it is at least orthogonal to the $n$ inliers $\bm X$. This motivates us to find such a basis $\bm q \in \mc S_{\mb X}^\perp$ by seeking a vector that is orthogonal to as many data points in $\mb Y$ as possible (i.e., the sparsest vector in $\bm Y^\perp$), resulting in \eqref{eqn:L1-ncvx} with $\varphi$ being the $\ell_1$-norm~\cite{manolis2015dualPCA,zhu2018dual,pmlr-v97-ding19b}. In this case, since the goal of DPCP is to compute a basis for ${\cal S}_{\mb X}^\perp$, the set of target solutions\footnote{Here, it should be noted that $\mc Q_\star$ not only contains the global minimizer, but also includes certain points that are not critical points \cite{tsakiris2018dual,zhu2018dual}.} is ${\cal Q}_\star = {\cal S}_{\mb X}^\perp \cap \bb S^{n-1}$. For the DPCP problem, it has proved in \cite{zhu2019linearly} that the optimization landscape \eqref{eqn:L1-ncvx} satisfies the regularity condition\footnote{This underlying regularity condition has been implicitly explored in \cite{manolis2015dualPCA,zhu2018dual} in convergence analysis.} \eqref{eq:RC-nonsmooth-0} with some positive $\alpha$ and sufficiently large $\epsilon$, where these parameters depend the dimension, size and distribution of the data points. Moreover, starting from a spectral initialization\footnote{For DPCP, we can compute such an initialization using the smallest eigenvector of $\mb Y\mb Y^\top$.} that falls in ${\cal B}({\cal Q}_\star,\epsilon)$, a basis vector for ${\cal S}_{\mb X}^\perp$ can be efficiently obtained by iterative algorithms which we will describe in more detail in \Cref{sec:optimization}. Similar local geometric properties have been generalized and studied in \cite{maunu2019well} for a new formulation that can lead to a direct estimation of the full basis for ${\cal S}_{\mb X}$.



Furthermore, this type of regularity condition also exist for other subspace models. For the orthogonal dictionary learning, results in \cite{bai2018subgradient,gilboa2019efficient} showed that a random initialization falls into regions satisfies 
\eqref{eq:RC-nonsmooth-0} with constant probability, ensuring fast convergence of gradient methods. This result is later extended to the sparse blind deconvolution problems \cite{qu2019blind,shi2019manifold}, where similar results are established. Finally, it should be noted that all these convergence guarantees are based on this underlying geometric property, that we will discuss in more detail about exploiting these properties for algorithmic design in \Cref{sec:optimization}.

\begin{figure}[t]
\centerline{
\begin{tikzpicture}
\node at (-2.5,0) {\includegraphics[width=2in]{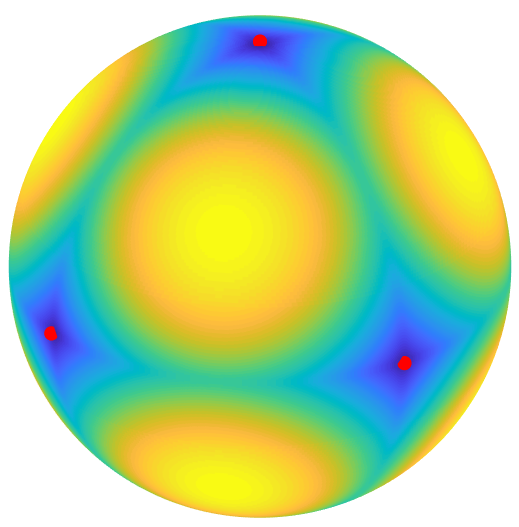}};
\node at (3,0) {\includegraphics[width=2in]{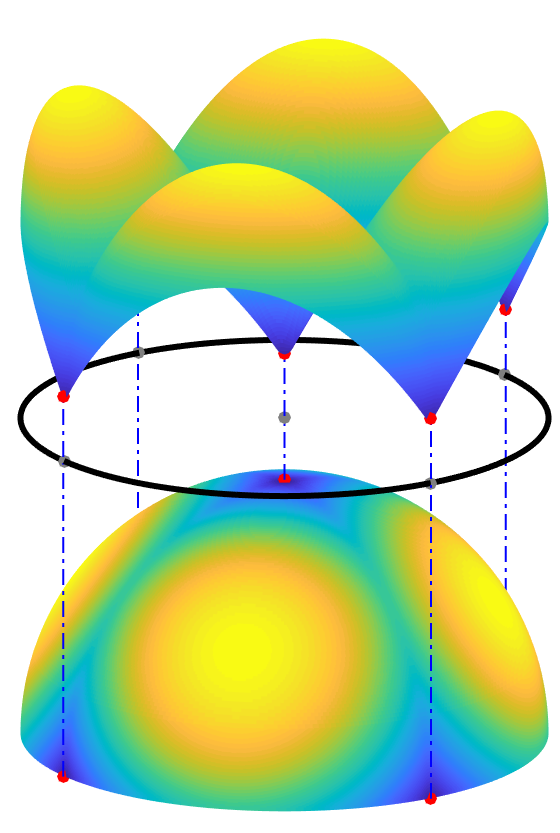}};
\end{tikzpicture}
}
\caption{{\bf Orthogonal DL with permutation symmetry.} Left: $ \varphi(\mb Y^\top \mb q)$ as a function on the sphere $\bb S^{2}$. Local minimizers (red) are signed standard basis vectors $\mc Q_\star = \Brac{\pm \mb e_i}$. These are the maximally sparse vectors on $\bb S^{2}$. Right: graph of $ \varphi(\mb Y^\top \mb q)$ reparameterized into Euclidean space; notice the strong negative curvature at points that are not sparse.} \label{fig:huber-sphere}
\end{figure}

\subsection{Global Geometry: Negative Curvature Near Saddles}\label{sec:global-geometry}

More surprisingly, recent work \cite{ge2015escaping,sun2016complete-i,sun2016geometric,li2018global,chi2018nonconvex} showed that in many cases nonconvex problems even have benign \emph{global} geometric structures (see \Cref{fig:huber-sphere} for an example), in the sense that
\begin{itemize}[leftmargin=*]
\item There is \emph{no} spurious local minimizer. (All) minimizers are (approximately) symmetric versions of the ground truth, and the optimization landscape around them exhibits local strong convexity or certain regularity properties that we discussed previously.
\item There is {\em no} flat saddle points. All saddles are created by symmetric superposition of the target solutions, and they exhibit negative curvature\footnote{Here, for $\mc C^2$ smooth functions, the negative curvature direction means the negative eigenvector direction of the Hessian.} in symmetry breaking directions.
\end{itemize}
These two characteristics circumvent two computational obstacles (see \Cref{fig:bad-landscape}) for nonconvex optimization: existences of (i) \emph{spurious} local minimizers and (ii) \emph{high-order} critical points. This implies that starting from any initialization, any optimization method which is able to efficiently escape saddle points converges to the global solution up to symmetry ambiguity. This type of function is also called \emph{strict saddle} functions \cite{ge2015escaping,sun2015nonconvex}.

\vspace{0.1in}
\noindent \textbf{Minimal Example II: Orthogonal Dictionary Learning.} In the context of finding the sparsest vector in a subspace $\mc S = \mathrm{row}(\mb Y)$, let us first use the orthogonal DL model $\mb Y = \mb A\mb X$ in \eqref{eq:DL Y} as an example to elaborate more on this type of global geometric structures. Recall from \Cref{sec:motivation}, given the generative model $\mb Y = \mb A \mb X \in \bb R^{n \times p} $ with orthogonal dictionary $\mb A \in \bb R^{n \times n} $ and sparse coefficient $\mb X \in \bb R^{n \times p}$, we aim to learn both $\mb A$ and $\mb X$ only given the data $\mb Y$. When the dictionary $\mb A$ is orthogonal, the observation is that $\mathrm{row}(\mb Y) = \mathrm{row}(\mb X)$ and the row vectors of $\mb X$ are sparse since $\mb X$ itself is sparse. When $\mb X$ is random and Bernoulli-Gaussian, the result in \cite{spielman2013exact} proved that the row vectors of $\mb X$ are the sparsest vectors in the subspace $\mc S = \mathrm{row}(\mb Y)$ provided $p \geq \Omega(n \log n)$. Therefore, we can reduce the orthogonal DL problem to finding one sparse row vector of $\mb X$ by solving the problem \eqref{eqn:L1-ncvx}. If one sparse row vector of $\mb X$ can be found, one may resort to deflation \cite{sun2016complete-i} or repeating random trials \cite{spielman2013exact,bai2018subgradient} to recover $\mb X$ and $\mb A$ up to a \emph{signed permutation} $\mathrm{SP}(n) $ ambiguity.

The reason that we can only solve the problem up to a $\mathrm{SP}(n) $ ambiguity, is because of the inherent symmetry structure, in the sense that signed permutation
\begin{align*}
    \mb Y \quad =\quad \mb A \mb X \quad = \quad  \Large( \mb A \mb \Gamma \Large)  \cdot \Large(  \mb \Gamma^\top \mb X \Large)
\end{align*}
creates \emph{equivalent feasible solutions}, where $\mb \Gamma \in \mathrm{SP}(n) $ is any signed permutation matrix. To see how this symmetry plays out in shaping the benign global optimization landscape, let us consider a simple case\footnote{For orthogonal $\mb A$, without loss of generality, we can always assume that $\mb A=\mb I$. This is because a change of variable $\ol{\mb q} = \mb A^\top \mb q $ reduces the problem \eqref{eqn:L1-ncvx} to the case $\mb A=\mb I$, which only rotates the optimization landscape.} that the dictionary $\mb A= \mb I$, so that $\mb Y = \mb X$ and the target solution set $\mc Q_\star$ of our optimization variable $\mb q$ becomes the set of signed standard basis vectors $\mc Q_\star = \Brac{ \pm \mb e_1, \cdots, \pm \mb e_n } $. Since the set $\mc Q_\star$ is also invariant to signed permutations, it is obvious that the function values $f( \mb \Gamma \mb q) = f(\mb q) $ for the problem \eqref{eqn:L1-ncvx}. As observed from \Cref{fig:huber-sphere}, for all critical points over the sphere: 
\begin{itemize}
    \item All the local minimizers are indeed globally optimal and close to the signed standard basis vector $\mb q_\star = \pm \mb e_i$ which is the ground truth. And the Riemannian Hessian at $\mb q_\star$ is positive definite tangent to $\mb q$, in the sense that
    \begin{align*}
         \Hess [f](\mb q_\star) \;=\; \mc P_{\mb q_\star^\perp} \paren{ \mb Y  \nabla^2 \varphi(\mb Y^\top \mb q_\star) \mb Y^\top  - \innerprod{ \mb Y \nabla \varphi(\mb Y^\top \mb q_\star )  }{ \mb q_\star } \mb I } \mc P_{\mb q_\star^\perp} \;\succeq\; \alpha \cdot \mc P_{\mb q_\star^\perp }
    \end{align*}
    for some $\alpha>0$, so that the function is strongly convex around $\mb q_\star$.
\item Saddle points $\mb q_s$ do exist, but they are balanced superpositions of target solutions
\begin{align*}
    \mb q_s \;=\; \frac{1}{\sqrt{ \abs{\mc I} } } \; \sum_{ i \in \mc I }\; \sigma_i \mb e_i,
\end{align*}
for every subset $\mc I \subseteq \Brac{ 1,\cdots, m } $ and sign scalar $\sigma_i \in \Brac{ \pm 1 } $. For each saddle point $\mb q_s$, the Riemannian Hessian manifests negative curvature, in the sense that
\begin{align*}
   \mb e_i^\top \Hess [f](\mb q_\star) \mb e_i \;<\; 0
\end{align*}
along the direction pointing to any $i$-th standard basis with $i \in \mc I$.
\end{itemize}
Since there is no spurious local minimizer presenting for the orthogonal DL, we can start from any point on the sphere and use any saddle point escaping method to find one sparse row vector from $\mb X$ via $\mb Y^\top \mb q_\star$. 

So far, we only considered a relative simple case in DL, where the dictionary $\mb A$ is orthogonal. If the dictionary is complete (i.e., square and invertible), we can approximately reduce the complete DL to orthogonal DL via simple techniques such as preconditioning (or whitening) of the data $\mb Y$ \cite{sun2016complete-i,zhai2019complete}. Aside from complete DL, recently similar benign global geometric properties have also been discovered for sparse blind deconvolution with multiple inputs \cite{li2018global} (see \Cref{fig:opt-landscape}). 
Similar to DL, this benign global landscape has also been induced by an intrinsic symmetry structure within the problem --- the shift symmetry. Indeed, every local minimizer for the sparse blind deconvolution is corresponding to a circulant shift of the unknown filter $\mb a_0$ in \eqref{eqn:mcs-bd-problem}. 

\Cref{tab:summary geometry} summarizes representative references on the local and global geometric properties for finding the sparsest vector in a subspace in the context of DPCP, DL, and MCS-BD that are illustrated in \Cref{sec:motivation}. Finally, we close this section by noting that the benign global geometric structure pertains to subspace models with certain symmetric structures, such as complete DL and sparse blind deconvolution. In both cases, the discrete symmetry such as permutation or shift only induce equivalent good solutions but no spurious local minimizers (see \Cref{fig:opt-landscape}). From this perspective, we conjecture that the DPCP problem could also obey benign global geometric property (by using a smooth objective). This is due to the fact that the \emph{continuous} symmetry such as rotations of the subspace may also only introduce equivalent good solutions but no spurious local minimizers.

 \begin{table*}  
\setlength{\tabcolsep}{11pt}
\renewcommand{\arraystretch}{1}
 \begin{center}
\resizebox{1\textwidth}{!}{
 \begin{tabular}{c|c|c|c|c}
 \hline
Objective $\varphi(\cdot)$ & Problem & \begin{tabular}{c} Distance between minimum\\ and the target solution\end{tabular}
 & Local geometry & Global geometry
 \\ \hline
 \multirow{2}{*}{\renewcommand{\arraystretch}{1}	$\ell^1$-norm}
& DL \cite{bai2018subgradient} &  0 & \color{green}\Large\checkmark &?\\ \cline{2-5}
   & DPCP \cite{tsakiris2018dual,zhu2018dual} & 0 & \color{green}\Large\checkmark & ?\\  \hline 
	Huber loss 
   & MCS-BD \cite{qu2019blind} & $\mc O(\mu)$ & \color{green}\Large\checkmark &?\\  \hline 
 \multirow{2}{*}{\renewcommand{\arraystretch}{1}	\begin{tabular}{c} $\mc C^\infty$ smooth \\ (e.g., Logcosh)\end{tabular}}
& DL \cite{sun2016complete-i,gilboa2019efficient} &  $\mc O(\mu)$ & \color{green}\Large\checkmark & \color{green}\Large\checkmark
   \\  \cline{2-5}
   & MCS-BD \cite{li2018global} & $\mc O(\mu)$ & \color{green}\Large\checkmark & \color{green}\Large\checkmark \\  \hline 
\end{tabular}
}
 \caption{\textbf{A selective summary of geometric analysis for problems \eqref{eqn:L1-ncvx}.} Here ? indicates that there is no existing result for this task, while $\color{green}\Large\checkmark$ denotes the existence of such a result.  }\label{tab:summary geometry}
 \end{center}
\end{table*}

\section{Efficient Nonconvex Optimization Methods}\label{sec:optimization}

The underlying benign geometric structures have strong implications for designing efficient, guaranteed optimization algorithms. In the following, we overview recent advances of optimization algorithms for solving problems \eqref{eqn:L1-ncvx} with different choices of convex surrogate $\varphi(\cdot)$ (see \Cref{tab:comparison-loss}), ranging from smooth to nonsmooth approaches, and from first-order methods to second-order methods to alternating minimization. For each method, we discuss the underlying principles and its advantages.

\subsection{Algorithms for smooth sparsity promoting convex surrogate $\varphi(\cdot)$}\label{subsec:optimization-smooth}

First, we consider the simplest setting where the function $\varphi(\cdot)$ in \eqref{eqn:L1-ncvx} is smooth. Undeniably, the most natural way to enforce sparsity is using nonsmooth surrogates such as $\ell^1$-penalty (e.g., $\varphi(\cdot) = \norm{ \cdot }{1} $). However, nonsmooth loss often results in substantial challenges in optimization and algorithmic analysis, due to the non-Lipschitzness of its subgradient. As shown in \Cref{tab:comparison-loss}, there are many ways to replace $\ell^1$-penalty with its smooth surrogate, where we can easily obtain benign global guarantees with relatively simple analysis. Nonetheless, the trade-off is that smoothing will introduce approximation errors; see \Cref{tab:summary geometry} for comprehensive summary. To have exact recovery, we need extra rounding step as shown in \cite{qu2016finding,sun2016complete-ii,qu2019blind}.

\paragraph{First-order methods.}
First, let us start with first-order iterative methods for solving the optimization problem \eqref{eqn:L1-ncvx} constrained over the sphere. As we explained in the last section that the Riemannian gradient describes the notion of slope over the sphere, one simple algorithm is to iteratively perform two steps --- move the iterate along the opposite direction of the Riemannian gradient and then project it back to the sphere --- which is known as Riemannian gradient descent (RGD) \cite{absil2009optimization}. In particular, in the $(k+1)$-th step, RGD updates the iterate by 
\begin{align}
\mb q^{(k+1)} \;=\; {\cal P}_{{\bb S}^{n-1}} \paren{\mb q^{(k)} - \eta_k \cdot \grad [f](\mb q^{(k)})},
\label{eq:RGD}\end{align}
where $\eta_k$ represents the step size which can be  chosen simply as a constant or selected by a Riemannian line search method \cite{absil2009optimization}. 

However, since RGD only uses the Riemannian gradient information, for general nonconvex problems it is only guaranteed to converge to a critical point \cite{absil2009optimization,boumal2018global}. In other words, it may get stuck at a saddle point because a critical point does not necessarily implies a local minimizer. Fortunately, for nonconvex problems with benign global geometry such as these considered in this paper, the work in  \cite{lee2019first} proved that RGD escapes from saddle points with negative curvature and converges to a second order critical point almost surely when using random initialization and constant step size. This escaping saddle property was also recently proved in \cite{panageas2019first} for RGD with varying step sizes. Therefore, when all saddle points exhibit negative curvature (or it is called \emph{strict saddle property} \cite{ge2015escaping}), RGD converges almost surely to a local minimum that satisfies second-order optimality condition. Furthermore, as we elaborated in \Cref{sec:geometry}, this type of local minimizer is close to a target solution for several important cases in finding the sparsest vector in a subspace.

Nonetheless, these results \cite{lee2019first,panageas2019first} do not directly imply how fast RGD escapes the saddle points and converges to a local minimum. In the worst case, the result in \cite{du2017gradient} showed that RGD can take exponential time converging to a local minimizer. This implies that the properties of strict saddle function are not sufficient for having polynomial convergence of RGD. For optimizing functions only having strict saddle property,
the only known result with global convergence rate is a perturbed version of RGD, which injects random noise into the descent iterates preventing stuck at saddle points. In particular, the results \cite{criscitiello2019efficiently,sun2019escaping} showed a sublinear convergence rate $\mc O(1/\epsilon^2)$ for noisy RGD, where $\epsilon$ is the accuracy tolerance\footnote{Precisely, it produces a point $\mb q$ with gradient smaller than $\epsilon$ and Hessian within $\sqrt{\epsilon}$ of being positive semidefinite, i.e., $\|\grad [f](\mb q)\|_2\le \epsilon, \Hess [f](\mb{q}) \succeq -\sqrt{\epsilon} \mc P_{\mb q^\perp}.$}. However, this type of results does not have direct implication in practice: (i) it is hard to control the level of noise to be injected, (ii) the convergence speed is hindered due to the randomly injected noises.

\begin{figure*}[t]
	\begin{subfigure}{0.3\linewidth}
		\centerline{
			\includegraphics[height=2in]{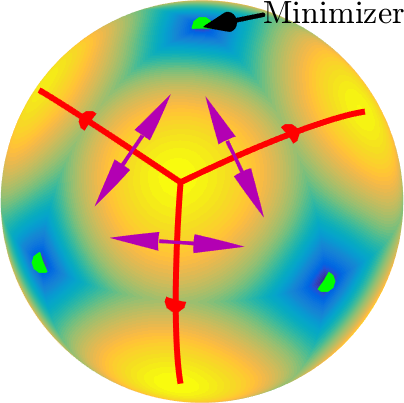}}
		\caption{ODL: Partition of the sphere \label{fig:odl-partition}}
	\end{subfigure}
	\hfill
	\begin{subfigure}{0.3\linewidth}
		\centerline{
			\includegraphics[height=2in]{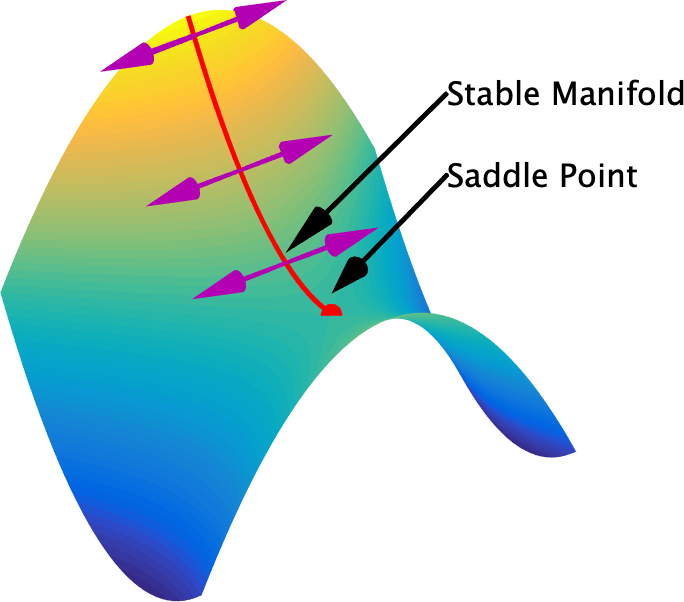}}
		\caption{ODL: Negative curvature \label{fig:odl-curvature}}
	\end{subfigure}
	\hfill
		\begin{subfigure}{0.3\linewidth}
		\centerline{
			\includegraphics[height=2in]{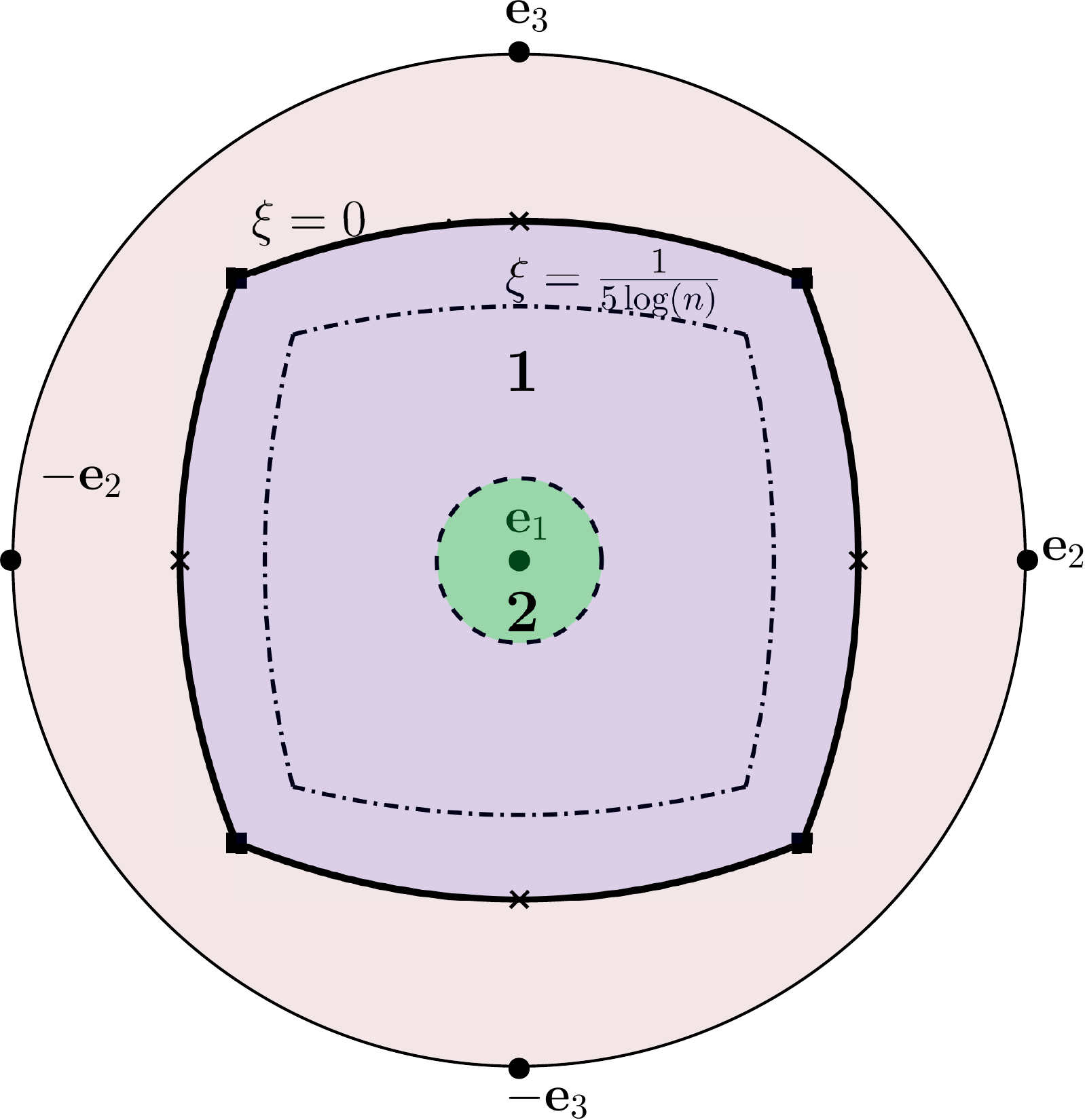}}
		\caption{ODL: large basin of attraction \label{fig:odl-set}}
	\end{subfigure}
	\caption{\small {\bf Illustration of partition of optimization landscape for orthogonal dictionary learning (ODL).} (a) shows a symmetric partition of the sphere based on the target solutions; (b) demonstrates large negative curvature along the stable manifold; (c) gives an illustration of the large basin of attraction around one target solution (the region inside the dotted line).}\label{fig:odl-gradient}
\end{figure*}

In practice, for many instances of finding the sparsest vector in a subspace, \emph{vanilla} RGD with random initializations seemingly always converge to the target solution with linear rate (see \Cref{fig:dpcp first order} and \Cref{fig:odl first order} for an illustration). This is mainly due to the fact that particular problems often have extra structures other than the strict saddle property. For example, in the orthgonal DL problem discussed in \Cref{sec:geometry}, as illustrated in \Cref{fig:odl-gradient}, the \emph{stable manifold} (i.e., the set of points along the flow that are sent towards the saddle point) exhibits strong negative curvature, that the gradient increases \emph{geometrically} moving away the the stable manifold (see \Cref{fig:odl-curvature}). Therefore, we have a large basin of attraction for each target solutions (see \Cref{fig:odl-set}), within which the function satisfies regularity conditions analogous to \eqref{eq:RC-nonsmooth-0} that we discussed in \Cref{sec:geometry}. Thus, it can be shown that a random initialization falls into one of these basins (the dotted region in \Cref{fig:odl-set}) with constant probability. In particular, when $\varphi(\cdot)$ is a $\log\cosh$ function, the result in \cite{gilboa2019efficient} rigorously showed this is true for orthogonal DL and proved sublinear convergence of RGD method. Similar ideas have been adopted for solving sparse blind deconvolution with multiple inputs \cite{qu2019blind} using a Huber loss, leading to an improved analysis that guarantees linear convergence of the vanilla RGD.

\begin{center}
\setlength{\arrayrulewidth}{0.4mm}
\setlength{\tabcolsep}{12pt}
\renewcommand{\arraystretch}{1.3}
 \begin{table*}  
 \resizebox{\textwidth}{!}{
 \begin{tabular}{c||c|c|c|c}
 \hline
Loss function $\varphi$ & Methods & Order  & \begin{tabular}{c} 
Convergence\\
(local)
\end{tabular} & \begin{tabular}{c} Complexity\\ (per iteration)\end{tabular}    \\ 
 \hline
\multirow{2}{*}{\renewcommand{\arraystretch}{1}	Smooth} & RGD &  $1$st & Linear & $\mc O(np)$  \\ 
\cline{2-5}
& RQN & $2$nd  & Quadratic & high \\
\hline
\multirow{3}{*}{\renewcommand{\arraystretch}{1}	Nonsmooth} & RSG &  $1$st & Linear & $\mc O(np)$   \\ 
\cline{2-5}
&  ManPPA & ?  & Quadratic & high (solving \eqref{eq:ManPPA}) \\
\cline{2-5}
&  IRLS & ?  & Linear & $\mc O(n^2p)$ \\
\hline
\end{tabular}
}
 \caption{\textbf{Summary of optimization methods for solving \eqref{eqn:L1-ncvx}.} Here ? means it is not clear the orders for ManPPA and IRLS algorihms. The computational complexity (per iteration) for RQN and ManPPA depends on the methods used for soving the subproblems, but in general their cost is much higher than merely  computing the gradient as in RGD or RSG.}\label{tab:comparison alg}
\end{table*}
\end{center}

\paragraph{Second-order methods.}

\begin{wrapfigure}{R}{0.37\textwidth}
\vspace{-.15in}
\centering	\includegraphics[height=1.5in]{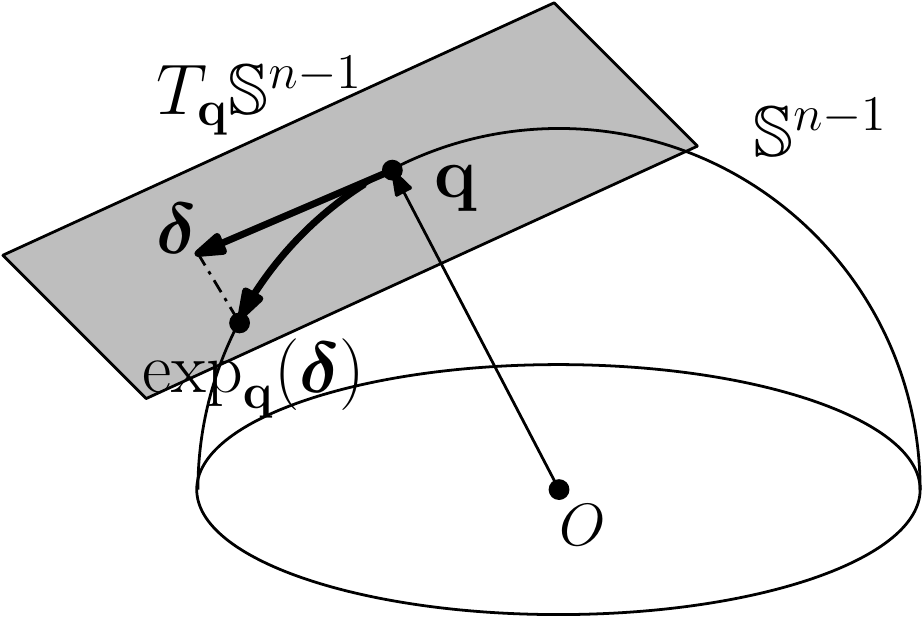}
	\caption{\textbf{Illustration of tangent space and retraction to the sphere.}
	}\label{fig:retraction}
\vspace{-.1in}
\end{wrapfigure}

Another important class of methods that can naturally escape strict saddle points are the second-order methods. This type of methods usually forms quadratic approximations of the function in the tangent space, and search for the descent direction based on this approximation within a restricted radius. At a saddle point $\mb q_s \in \bb S^{n-1}$ with $\grad [f](\mb q_s) = \mb 0 $, this type of methods can directly exploit the Hessian information to find the descent direction that is aligned with the negative eigenvector of the Hessian. To see why this happens, consider the following quadratic approximation $\widehat{f}(\cdot)$ of the function $f(\cdot)$ at a saddle point $\mb q_s \in \bb S^{n-1}$, 
\[
\widehat{f}\paren{ \mb q_s + \mb d } \;=\; f(\bm q_s) +  \frac{1}{2}\mb d^\top \Hess [f](\mb q_s) \mb d + \mc O(\norm{\mb d}{2}^3),  \;\; \forall\; \mb d\in \mathrm{T}_{\mb q_s}\bb S^{n-1}.
\]
Because the Riemannian Hessian at $
\mb q_s$ has negative eigenvalues, if the direction $\mb d$ is aligned with the negative eigenvector of $\Hess [f](\mb q_s)$, then we have $\mb d^\top \Hess [f](\mb q_s) \mb d <0$. Since  $\widehat{f}\paren{ \mb q_s + \mb d }\approx f\paren{ \mb q_s + \mb d }$ for small $\norm{\mb d}{2}$, this further implies that $f\paren{ \mb q_s + \mb d }<f\paren{ \mb q_s }$.  In other words, the strict saddle points can be efficiently escaped with second-order methods, directly exploiting the negative curvature information of the Riemannian Hessian.

In addition, when optimizing over the sphere, it should be noted that the Riemannian second-order methods here can be viewed as a natural extension of classical second-order methods in the Euclidean space. The quadratic approximation is formed using Riemannian derivatives in the tangent space (which is also a linear space), while the only difference is that we need to perform an extra retraction step to retract the iterate from tangent space back to the sphere (see \Cref{fig:retraction}). These Riemannian second-order methods include Riemannian Newton method \cite{absil2009optimization}, and Riemannian Quasi-Newton (RQN) method (e.g., the Riemannian trust-region method \cite{absil2009optimization} and the Riemannian cubic-regularization method \cite{zhang2018cubic}). We omit the algorithmic details and refer interested readers to the references \cite{absil2009optimization,zhang2018cubic} for a closer look.

In comparison with first-order methods, the major advantage of second order methods is the convergence speed. As can be seen from \Cref{fig:dpcp experiments} and \Cref{fig:odl experiments}, the second methods are approximately 10 times faster than first-order methods in terms of iteration complexity. In theory, for example, the Riemannian trust-region method is proved to converge to a target solution at a local quadratic rate for complete DL \cite{sun2016complete-i}. Nonetheless, for each iteration the computation and memory costs of the second methods are usually much higher than first-order methods, which is due to the fact that they need to solve an expensive subproblem.
Therefore, it is often preferred to use second-order methods for small-scale problems, and use first-order methods for large-scale ones. We summarize the comparison of algorithms in \Cref{tab:comparison alg}.
As a future work, it is interesting to design algorithms with similar fast convergence but much lower computation cost per iteration (e.g., Riemannian versions of limited-memory BFGS algorithms \cite{liu1989limited,yuan2016riemannian}).

\subsection{Algorithms for nonsmooth sparsity promoting convex surrogate $\varphi(\cdot)$}\label{subsec:optimization-nonsmooth}

\begin{figure*}[t]
	\begin{subfigure}{0.48\linewidth}
		\centerline{
			\includegraphics[width=3.2in]{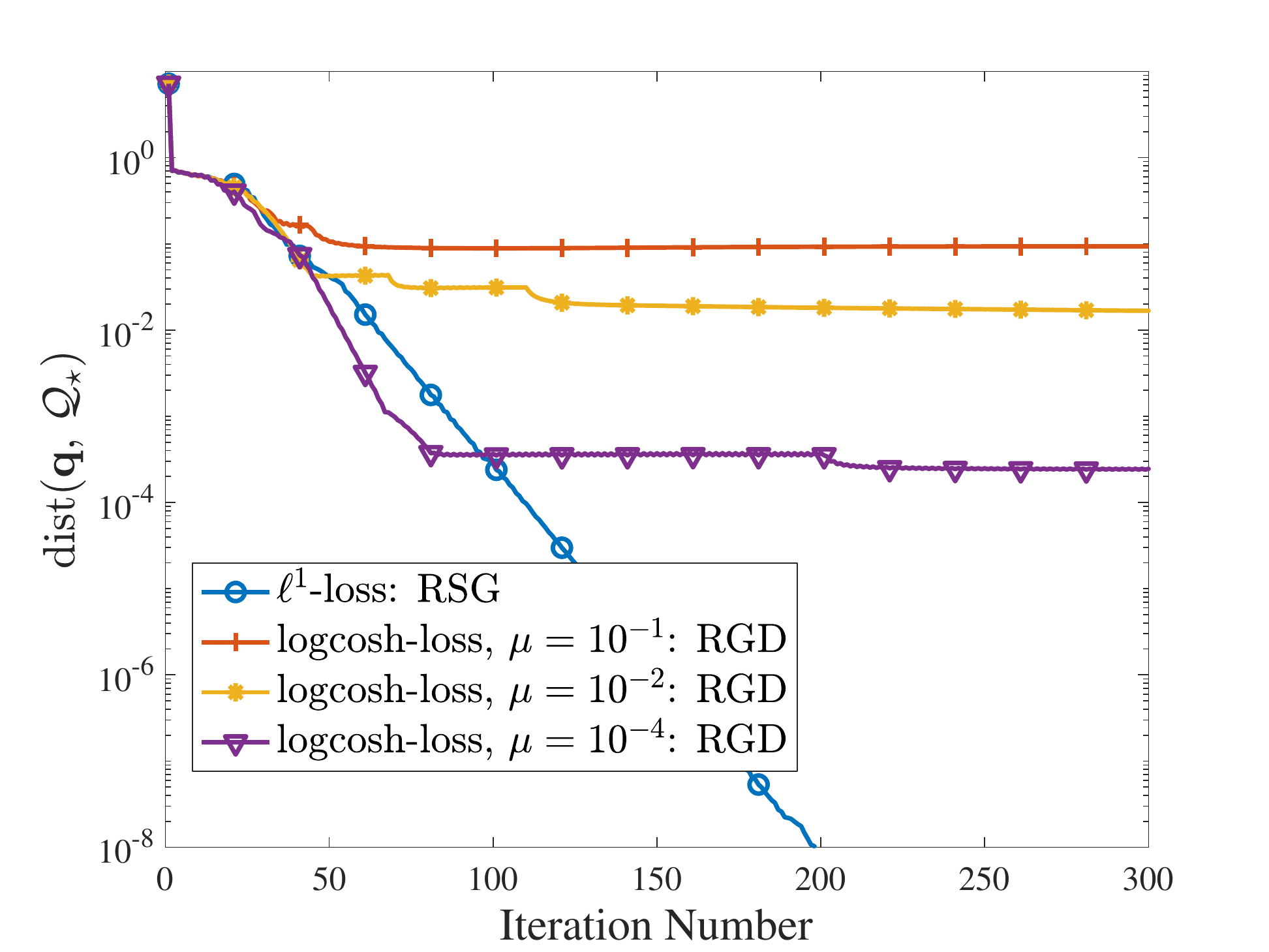}}
		\caption{DPCP: First-order methods \label{fig:dpcp first order}}
	\end{subfigure}
	\begin{subfigure}{0.48\linewidth}
		\centerline{
			\includegraphics[width=3.2in]{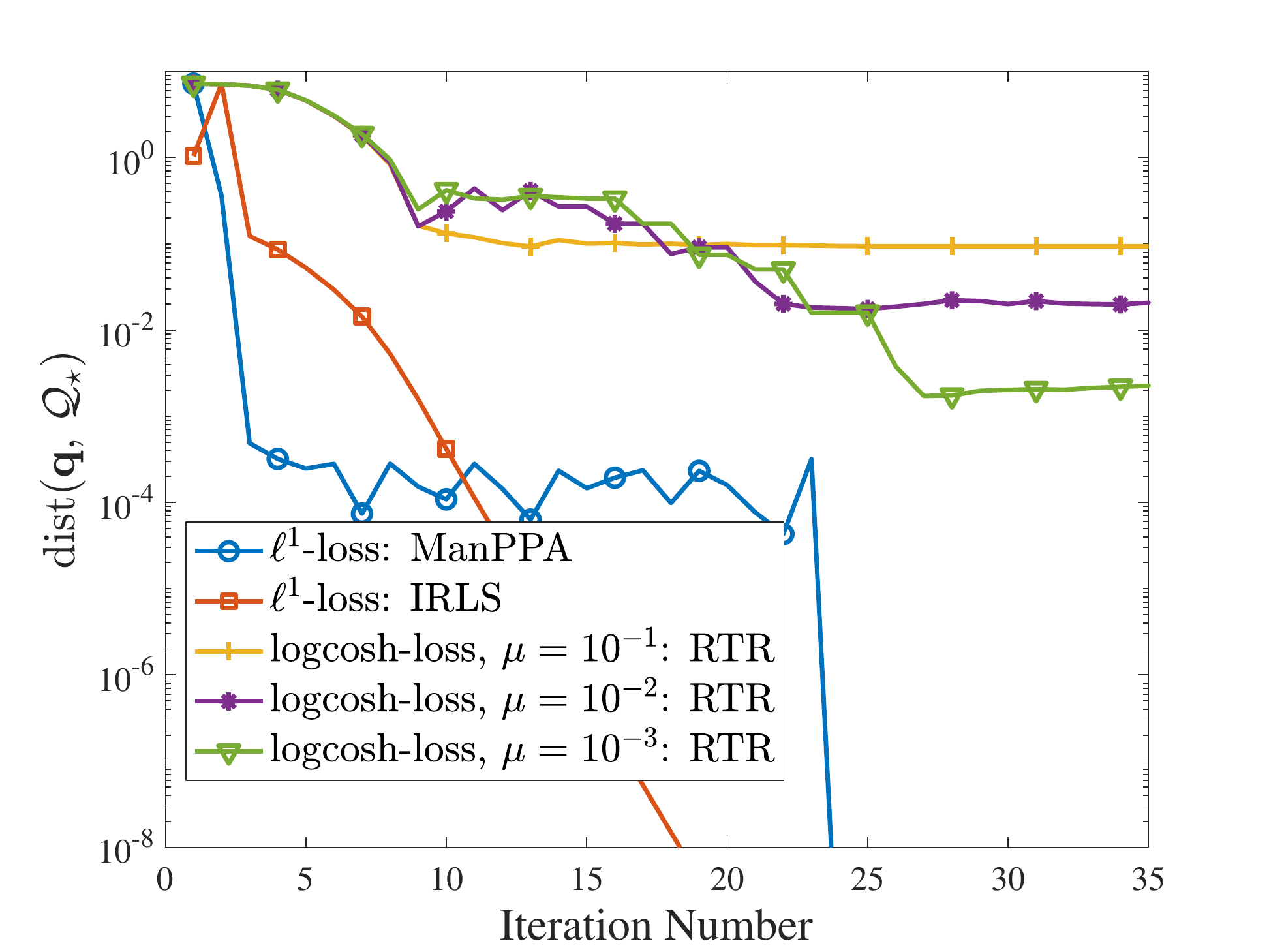}}
		\caption{DPCP: High-order methods \label{fig:dpcp second order}}
	\end{subfigure}
	\caption{\small {\bf Comparison of convergence performances for the DPCP problem.} As illustrated in \eqref{eq:DPCP Y} of \Cref{sec:motivation}, we generate the data similar to \cite{zhu2018dual}, where the subspace ${\cal S}_{\mb X}$ is randomly sampled  with co-dimension $r=60$ with ambient dimension $n = 100$. We then generate $p_1 = 1500$  inliers uniformly at random from the unit sphere in  ${\cal S}_{\mb X}$ and $p_2 = 3500$ outliers uniformly at random from the unit sphere in $\R^n$. We initialize all the algorithms at the same point with its entries follow i.i.d. standard Gaussian distribution. We examine performances of the optimization algorithms on two loss functions: (i) $\mc C^\infty$-smooth loss $\varphi(\cdot) = \mu \log \cosh (\cdot/\mu )$ and (ii) nonsmooth $\varphi(\cdot) = \norm{ \cdot }{1}$. For the smooth loss, we test first-order RGD and the second-order Riemannian trust-region (RTR) method introduced in \Cref{subsec:optimization-smooth}. For the nonsmooth loss, we test first-order RSG method, and high-order ManPPA and IRLS methods introduced in \Cref{subsec:optimization-nonsmooth}. \Cref{fig:dpcp first order} illustrates convergence performances of first order methods; \Cref{fig:dpcp second order} illustrates convergence performances of high order methods. }\label{fig:dpcp experiments}
\end{figure*}

\begin{figure*}[t]
	\begin{subfigure}{0.48\linewidth}
		\centerline{
			\includegraphics[width=3.2in]{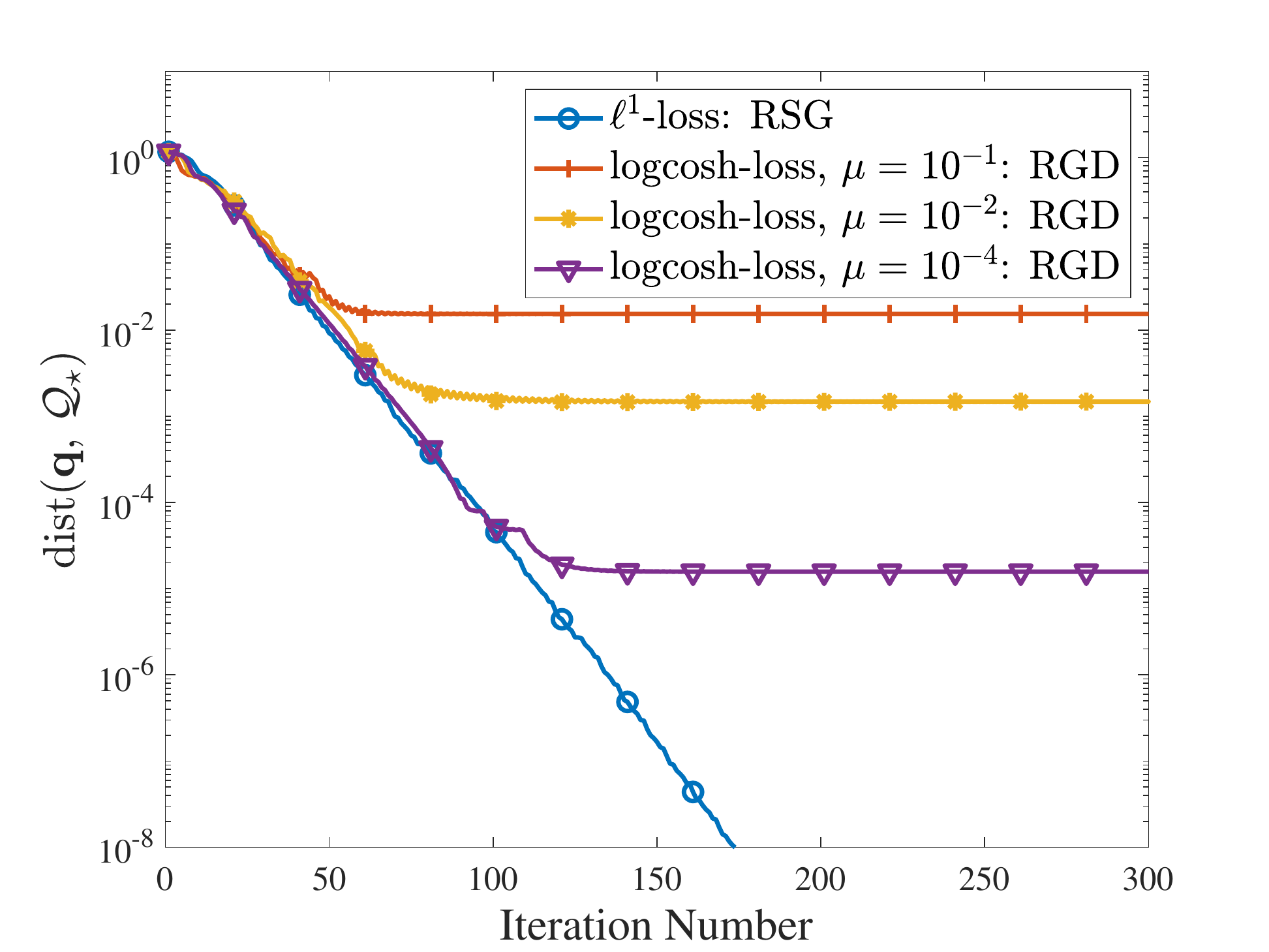}}
		\caption{ODL: First-order algorithms \label{fig:odl first order}}
	\end{subfigure}
	\begin{subfigure}{0.48\linewidth}
		\centerline{
			\includegraphics[width=3.2in]{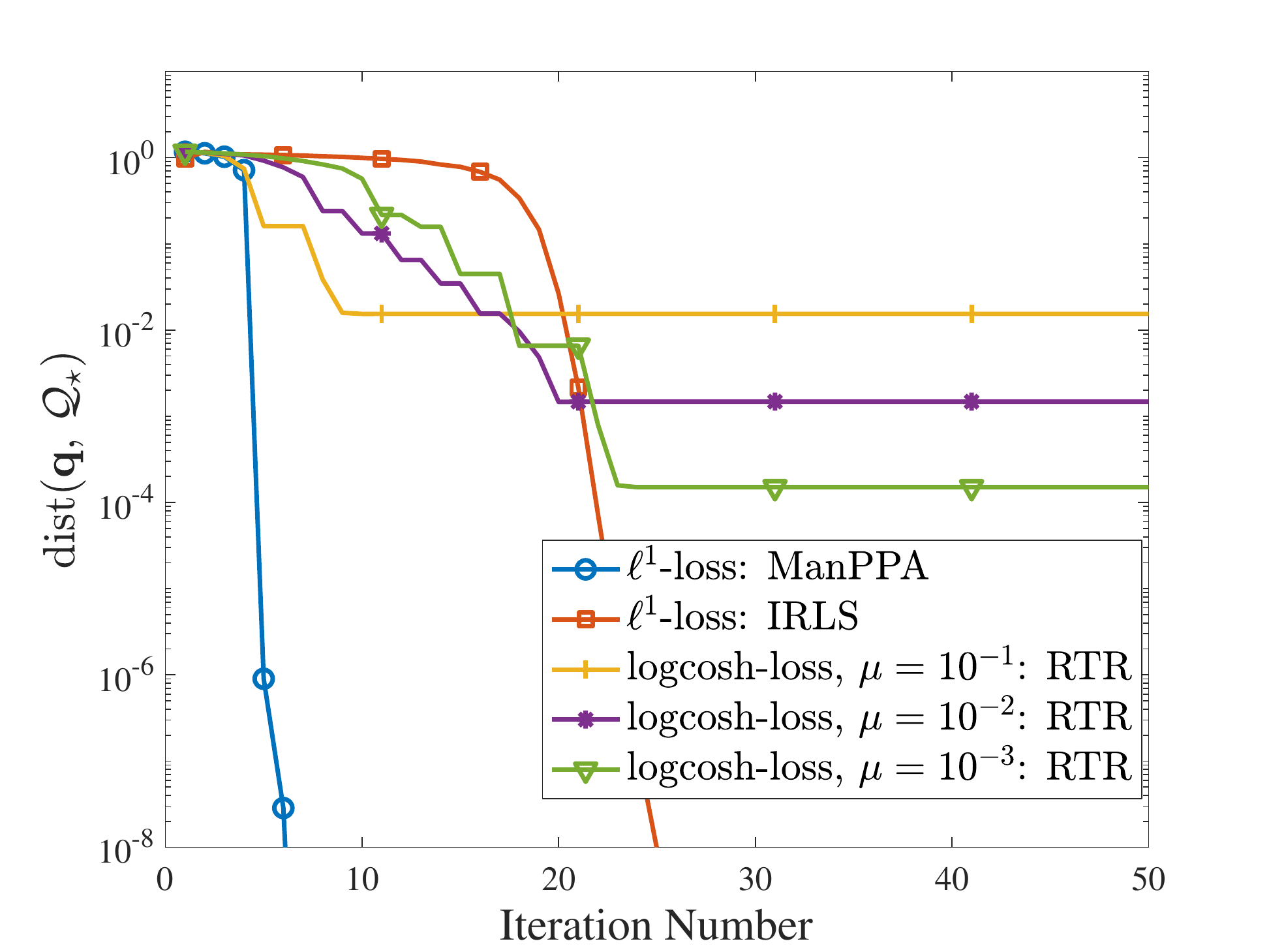}}
		\caption{ODL: High-order algorithms \label{fig:odl second order}}
	\end{subfigure}
	\caption{\small {\bf Comparison of convergence performances for the ODL problem.} As illustrated in \eqref{eq:DL Y} of \Cref{sec:motivation}, we generate the data $\mb Y=\mb A \mb X$ similar to \cite{bai2018subgradient}, where we create the dictionary $\mb A \in \bb R^{n\times n}$ as a random orthogonal matrix with $n =64$, and draw each entry of the sparse code $\mb X \in \bb R^{n \times p} $ from i.i.d. Bernoulli-Gaussian distribution with sparsity level $\theta= 0.25$ and $p\approx 10 n^{1.5}$. We initialize all the algorithms at the same point with its entries follow i.i.d. standard Gaussian distribution. We examine performances of the optimization algorithms on two loss functions: (i) $\mc C^\infty$-smooth loss $\varphi(\cdot) = \mu \log \cosh (\cdot/\mu )$ and (ii) nonsmooth $\varphi(\cdot) = \norm{ \cdot }{1}$. For the smooth loss, we test first-order RGD and the second-order Riemannian trust-region (RTR) method introduced in \Cref{subsec:optimization-smooth}. For the nonsmooth loss, we test first-order RSG method, and high-order ManPPA and IRLS methods introduced in \Cref{subsec:optimization-nonsmooth}. \Cref{fig:odl first order} illustrates convergence performances of first order methods; \Cref{fig:odl second order} illustrates convergence performances of high order methods. }\label{fig:odl experiments}
\end{figure*}

As alluded in \Cref{subsec:optimization-smooth} (see \Cref{tab:summary geometry}), optimizing smooth surrogates often induces approximation errors of the solution, so that extra steps are required for finding the exact target solutions. In contrast, optimizing nonsmooth objectives directly produces exact solutions, demonstrated in \Cref{fig:dpcp experiments} and \Cref{fig:odl experiments}. In the following, we review recent advances on developing nonsmooth optimization methods for finding the sparsest vector in a subspace.

\paragraph{Riemannian SubGradient (RSG) methods.}

A natural modification of RGD for a non-smooth $\varphi(\cdot)$ (i.e., the $\ell^1$-loss) is a Riemannian SubGradient (RSG) method  that replaces the Riemannian gradient by a Riemannian subgradient in \eqref{eq:RGD}. Although each iterate of RSG has a similar form as RGD in \eqref{eq:RGD}
\begin{align}
\mb q^{(k+1)} \;=\; {\cal P}_{{\bb S}^{n-1}} \paren{\mb q^{(k)} - \eta_k \cdot \partial_R f(\mb q^{(k)})},
\label{eq:RSG}\end{align}
the convergence behavior of RSG is much more complicated than RGD because of the nonsmoothness. For example, unlike RGD, the negative Riemannian subgradient $-\partial_R f(\bm q)$ is not necessarily a descent direction\footnote{How to efficiently search for an appropriate descent direction to accelerate the convergence for nonsmooth objective is still an open and interesting question. Existing methods such as Riemannian gradient sampling algorithm \cite{hosseini2017riemannian} is often very expensive and lacks non-asymptotic convergence guarantees.}, and the RSG with a constant step size may even fail to converge to a critical point\footnote{This is true even when there is no sphere constraint~\cite{Shor85}. As a simple example, consider minimizing $|x|$ by the subgradiment method $x_{k+1} = x_k - \eta_k \sign(x_k)$ and suppose that we take $x_0=0.01$ and $\eta_k = 0.02$ for all $k\ge0$. Then, the iterates $\{x_k\}_{k\ge0}$ will oscillate between the two points $x_+=0.01$ and $x_-=-0.01$ and never converge to the global minimum $x^\star=0$. At best, one can only show that even for a convex function, the subgradient method with a constant step size will converge to a neighborhood of the set of global optima (with rate guarantees if the problem satisfies additional regularity conditions); see, e.g.,~\cite{Shor85,NB01,B12,davis2018subgradient}. To ensure the convergence of subgradient methods, a set of diminishing step sizes is generally needed~\cite{Shor85,goffin1977convergence}.}. Moreover, the convergence analysis of RSG for general nonsmooth Riemannian optimization problems are still largely unexplored \cite{absil2019collection}. It is only very recently that \cite{li2019nonsmooth} provided the first convergence rate guarantees for RSG for optimizing nonsmooth functions over Stiefel manifold (which includes the sphere as a special case), under certain regularity conditions of the function. More specifically, if the objective function is \emph{weakly convex}\footnote{We say $f$ is weakly convex if there exists a $\tau$ such that $f(\cdot) + \frac{\tau}{2}\norm{\cdot}{2}^2$ is convex.} in the Euclidean space, then RSG with an arbitrary initialization and diminishing step size (e.g., $\eta_k = 1/\sqrt{k}$) converges to a critical point at a sublinear rate (e.g., $\mc O(1/k^{1/4})$) \cite{li2019nonsmooth}.

Moreover, when the nonsmooth objective satisfies the local regularity condition\footnote{ \cite{li2019nonsmooth} utilizes another property called sharpness, which together with the weak convexity also results a similar regularity condition \eqref{eq:RC-nonsmooth-0}.} \eqref{eq:RC-nonsmooth-0}, very recent results \cite{zhu2018dual,maunu2019well} showed that RSG with (piecewise) \emph{geometrically shrinking} step size converges with a linear rate. This type of results is quite \emph{surprising}, in the sense that our common knowledge tell us that RSG is usually the slowest method for optimizing nonsmooth functions. Nonetheless, the key elements for enjoying fast linear convergence using RSG are: {\em (i)} the underlying benign geometric structure of the problem --- the local regularity condition \eqref{eq:RC-nonsmooth-0}, and {\em (ii)} the use of geometrically diminishing stepsize (i.e., $\eta_k = \mc O(\beta^k)$ for some properly chosen $\beta\in(0,1)$). Thus, for problems such as DPCP and ODL, once we initialize within the local region ${\cal B}({\cal Q}_\star,\epsilon)$ around the target solution $\mc Q_\star$, then RSG converges to $\cal Q_\star$ with a linear rate \cite{zhu2019linearly}, i.e., $\dist(\mb q^{(k)}, {\cal Q}_\star)\lesssim \beta^k$. To avoiding tuning the step size, recently a modified  backtracking line search technique can be used to automatically search an appropriate step size \cite{nocedal2006numerical,zhu2018dual}.

\paragraph{Manifold proximal point algorithm (ManPPA).} The manifold proximal point algorithm (ManPPA) \cite{chen2019manifold}, which adopts the idea of the classical proximal point method in the Euclidean space,  is an effective approach to find a descent direction on the Moreau envolope of $\varphi(\cdot)$ within the tangent space:
\begin{equation}\begin{split}
    \bm d^{(k)} \;&=\;\argmin_{\bm d \in \R^{n}}\; \varphi\paren{\bm Y^\top (\bm q^{(k)} + \bm d)} + \frac{1}{2t_k}\norm{\bm d}{2}^2  \quad \text{s.t.} \;\; \bm d^\top \bm q^{(k)} \;=\; 0, \\  \bm q^{(k+1)} \;&=\; {\cal P}_{{\bb S}^{n-1}} \paren{\bm q^{(k)} + \alpha_k\cdot \bm d^{(k)} },
\end{split}
\label{eq:ManPPA}\end{equation}
where $t_k>0$ and $\alpha_k>0$ are the step sizes. The efficiency of ManPPA depends on whether we can efficiently solve the optimization subproblem for the descent direction in \eqref{eq:ManPPA}. The work in \cite{chen2019manifold} solves this convex subproblem by using an inexact augmented Lagrangian method together with a semi-smooth Newton method. In comparison with RSG, ManPPA converges much faster in terms of iteration complexity\footnote{A local quadratic convergence rate is established in \cite{chen2019manifold} for problems obeying shaprness, which is satisfied for both DPCP and the orthogonal dictionary learning \cite{li2019nonsmooth}.}, but its overall computational complexity can still be higher because solving the subproblem in \eqref{eq:ManPPA} is usually quite expensive even with efficient implementations.

\paragraph{Alternating linearization and projection (ALP) method.} 
Another way to deal with nonsmooth $\ell^1$-minimization problem with nonlinear constraint is simply to linearize the constraint via linear approximations, and solve a sequence of linear programs (LPs) until convergence. This is the so-called alternating linearization and projection (ALP) method \cite{spath1987orthogonal,manolis2015dualPCA}. In particular, for our problem \eqref{eqn:L1-ncvx}, we linearize the spherical constraint $\mb q^\top \mb q = 1$ by using its first order Taylor approximation at the point $\mb q^{(k)}$, resulting in a linear constraint $\bm q^\top \bm q^{(k)} =1$. Thus, we compute a sequence of iterates ${\bm q^{(k)}}$ via solving the following subproblem
\begin{equation}
    \overline{\bm q}^{(k)} \;=\; \argmin_{\bm q\in \R^{n}} \norm{\bm Y^\top \bm q}{1}, \quad  \text{s.t.} \quad  \bm q^\top \bm q^{(k)} \;=\; 1,  \quad \text{and} \quad  \bm q^{(k+1)}\; =\; {\cal P}_{{\bb S}^{n-1}} \paren{ \overline{\bm q}^{(k)} },
\label{eq:alg ALP}\end{equation}
where the optimization subproblem is simply an LP. It turns out that ALP can be viewed as a special instance of ManPPA by choosing $t= \infty$ and $\alpha_k = 1$ in \eqref{eq:ManPPA} and setting $\mb q = \mb q^{(k)} + \mb d$ in \eqref{eq:alg ALP}.

For general nonconvex problems, Sp\"{a}th and Watson \cite{spath1987orthogonal} established the convergence of ALP to a critical point. For the DPCP problem, this proving technique is further utilized in \cite{manolis2015dualPCA,zhu2018dualArxiv} to show the convergence to a target solution starting from a spectral initialization. Again, the latter result is achieved mainly due to the underlying benign geometric structures of the problem that we discussed in  \Cref{subsec:local-geometry}. In practice, the ALP usually converges much faster than RSG in terms of iteration complexity. However, since for each iteration it involves solving an LP (e.g., can be solved using Gurobi \cite{optimization2014inc}) time consuming subproblem, the overall computational complexity could still be high.

Finally, we note that if our initial point $\bm q^{(0)}$ is very close to a global minimizer, solving one LP in \eqref{eq:alg ALP} exactly returns the target solution. This property has been explored in \cite{qu2014finding,sun2016complete-i,qu2019blind} for rounding approximate solutions (often produced by optimizing smooth objectives) to the exact target points. Moreover, for the rounding step the work in \cite{qu2019blind} proposed an efficient projected subgradient method that enjoys local linear convergence.

\paragraph{Iterative reweighted least squares (IRLS).} While the ALP iteratively linearizes the nonconvex constraint, the iterative reweighted least squares (IRLS) \cite{manolis2015dualPCA,lerman2015robust,lerman2017fast,zhang2014novel} attempts to smooth the nonsmooth objective by a weighted least squares. It should be noted that the IRLS is a classical method to solve $\ell^p$-minimization problems ($p\neq 2$) such as compressive sensing~\cite{candes2008enhancing,chartrand2008iteratively,daubechies2010iteratively}. The main idea behind IRLS is to alternatively solve a weighted least-squares problem (which often admits a closed-form solution) and update the weights. To illustrate the IRLS for solving \eqref{eqn:L1-ncvx} \cite{manolis2015dualPCA,lerman2015robust,lerman2017fast,zhang2014novel}, let us consider $\varphi(\cdot) = \norm{\cdot}{1}$ and rewrite $\varphi(\mb Y^\top \mb q)$  as $\|\bm Y^\top \bm q\|_1 = \sum_{i=1}^p \abs{\bm y_i^\top \bm q} = \sum_{i=1}^p \frac{1}{\abs{\bm y_i^\top \bm q}} (\bm y_i^\top \bm q)^2$. This inpsires us to consider solving the following subproblem
\begin{equation}\label{eqn:subproblem-irls}
\bm q^{(k)} = \argmin_{\bm q\in \bb S^{n-1}}\sum_{i=1}^p w_{i}^{(k-1)} (\bm y_i^\top \bm q)^2, \quad \text{and} \quad w_{i}^{(k)} = \frac{1}{ \max\{\delta, \abs{\bm y_i^\top \bm q^{(k)} } \} } \ \forall i\in[p],
\end{equation}
where $\delta$ is a small scalar to avoid numerical explosion. It is not difficult to show that the optimal solution of the subproblem \eqref{eqn:subproblem-irls} is given by the eigenvector corresponding to the smallest eigenvalue of $\sum_{i=1}^p w_{i}^{(k-1)}  \bm y_i\bm y_i^\top$. The convergence behavior of IRLS is discussed in \cite{lerman2017fast}, where the global convergence to a critical point and a local convergence to an approximate target solution is established for solving DPCP. In comparison to RSG, IRLS converges much faster and it does not require tuning the step size (see \Cref{fig:dpcp experiments} and \Cref{fig:odl experiments}). However, similar to ALP, the subproblem of IRLS is expensive as it requires performing an eigen-decomposition.

\paragraph{Other methods.}
Finally, we close this section by noting that there are many other methods developed for constrained nonsmooth problems that may also be used for solving \eqref{eqn:L1-ncvx}. Typical examples include \cite{curtis2012sequential} SQP-GS (which combines sequential quadratic programming (SQP) and gradient sampling (GS) techniques), and a faster quasi-Newton type method which called GRANSO \cite{curtis2017bfgs} 
which improves SQP-GS by employing the BFGS method. GRANSO has been used for solving orthogonal DL in \cite{bai2018subgradient} and converges very fast in practice, but there has been no convergence guarantee established yet.

\section{Applications in learning low-complexity models from the data}\label{sec:application}

\begin{figure*}[t]
	\centering
	\centering
    \begin{minipage}[c]{0.3\textwidth}
    	\centering
    	\includegraphics[width = 0.9\linewidth]{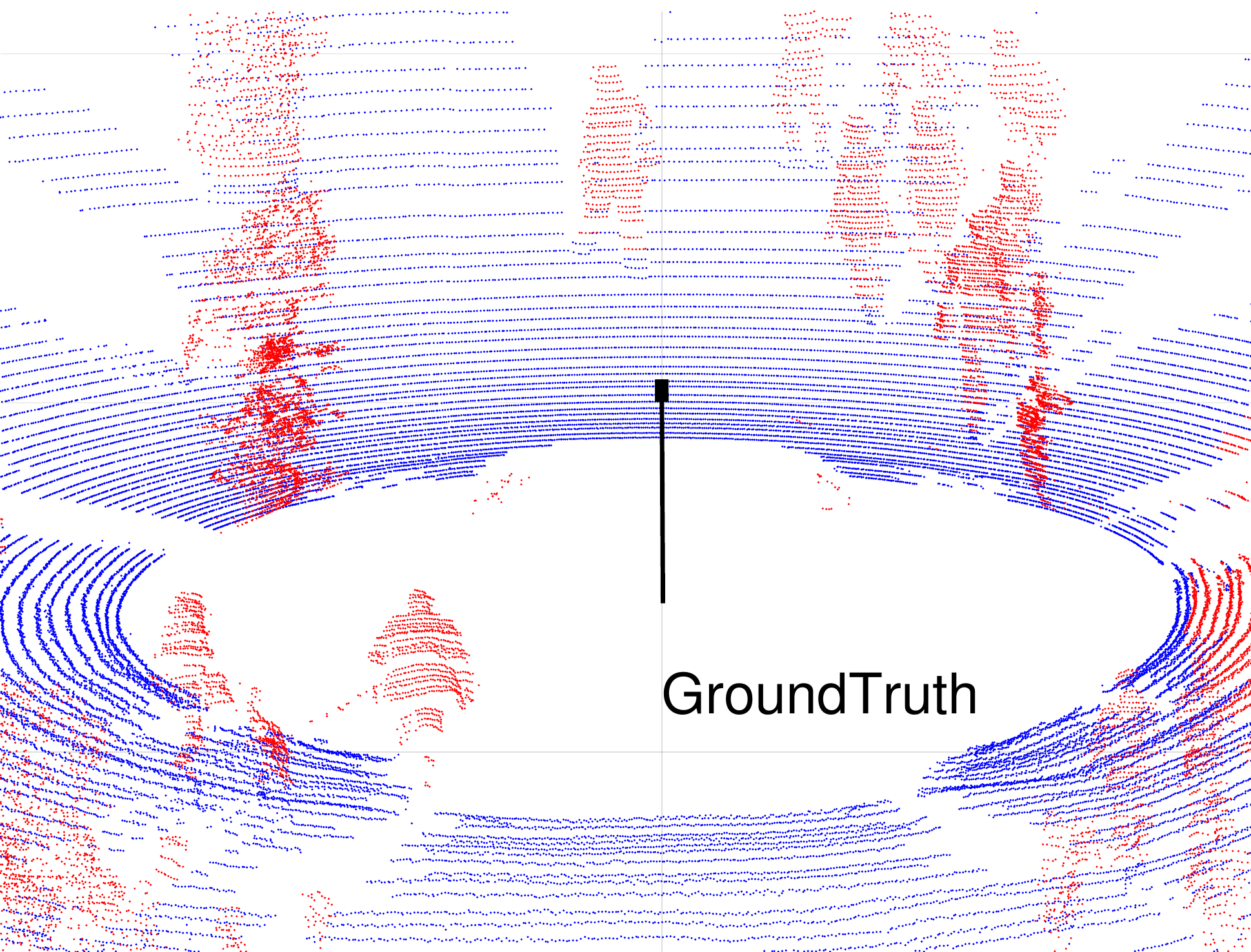}
    	\subcaption{Annotation}
    \end{minipage}
    \begin{minipage}[c]{0.3\textwidth}
    	\centering
    	\includegraphics[width = 0.9\linewidth]{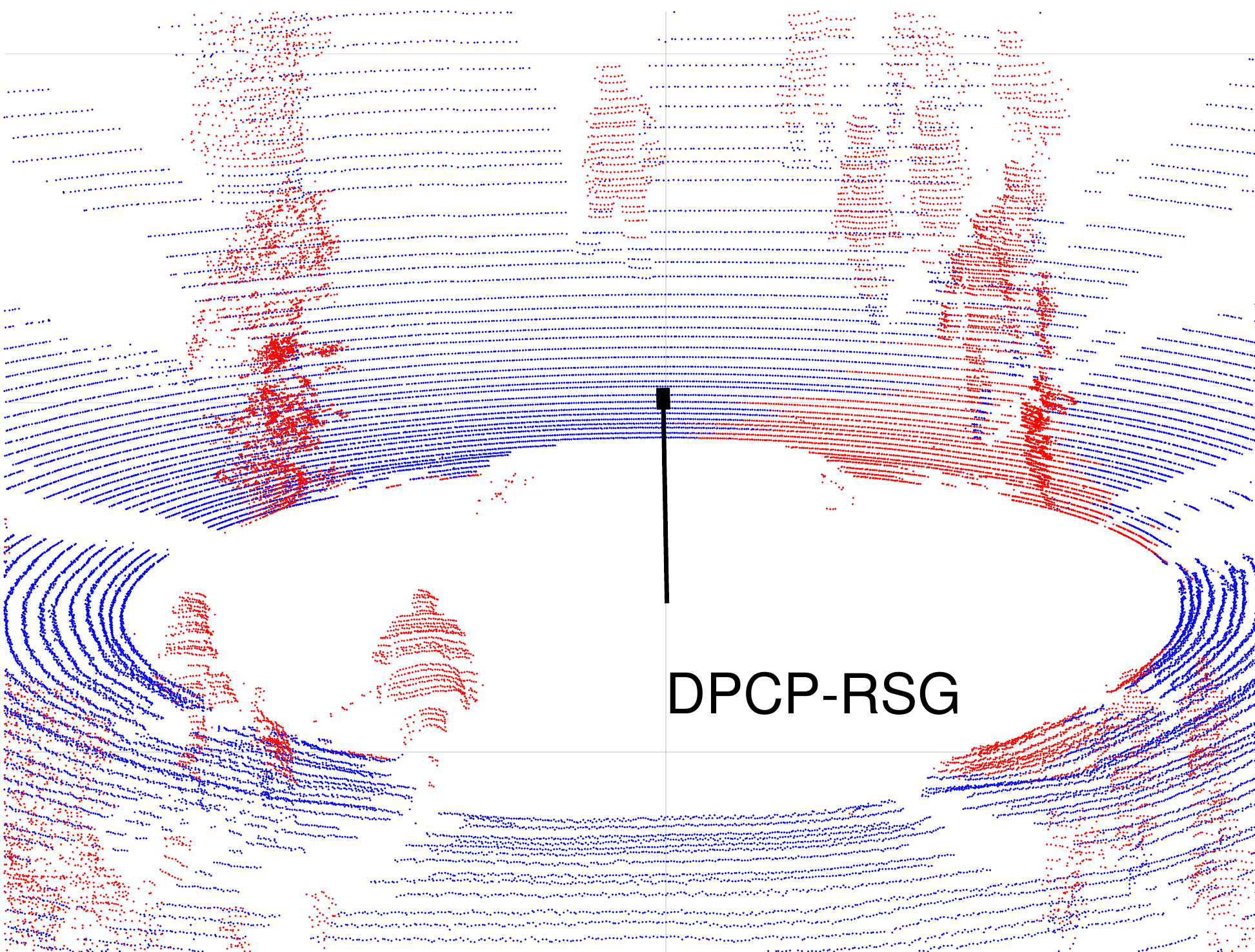}
    	\subcaption{DPCP approach}
    \end{minipage}
    \begin{minipage}[c]{0.36\textwidth}
    	\centering
    	\includegraphics[width = 0.9\linewidth]{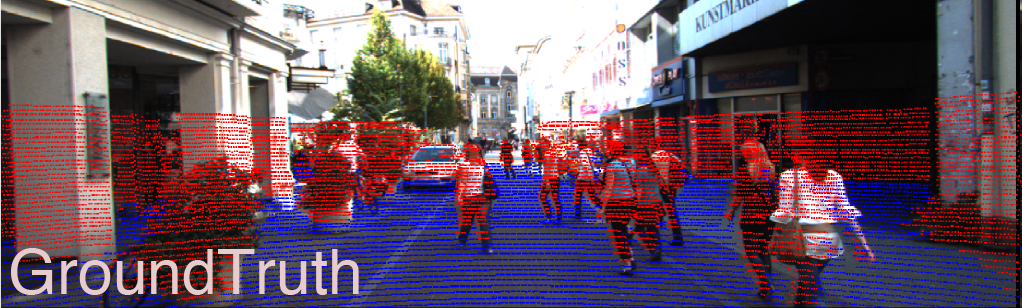}\\
    	\vspace{.1in}
    	\includegraphics[width = 0.9\linewidth]{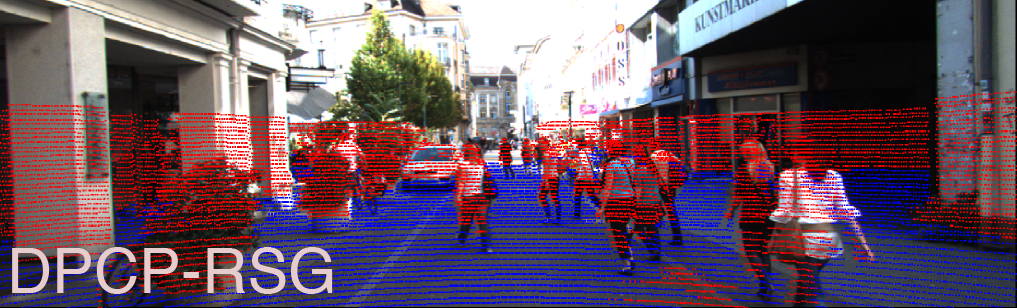}
    	\subcaption{Projection onto image}
    \end{minipage} 
	\caption{\textbf{Road detection for autonomous driving \cite{zhu2018dual,ding2019noisy}.} Illustration of results on Frame 328 of KITTY-CITY-71 \cite{geiger2013vision}: (a) annotated 3D point clouds where inliers (on the road plane) are indicated by blue and outliers (off that plane) are indicated by red, (b) DPCP approach using the RSG method, (c) projections of 3D point clouds onto the image.}
	\label{fig:app-dpcp}
\end{figure*}

High dimensional data often possess low dimensional structures such as sparsity. For a variety of applications in data science, one of the fundamental problems that we are facing today is how to learn those low-complexity structures/models only given the data. In the following, we present several engineering applications for which some of these challenging learning problems can be reduced to the task of finding the sparsest vector in a subspace. Therefore, we can leverage on the nonconvex optimization approaches illustrated in this work, efficiently solving these problems with global provable guarantees. 

\paragraph{Machine Intelligence.} In many applications such as national security, autonomous driving, healthcare, we want to endow our system the ability to correctly interpret external data, to learn from such data, and to use those learnings to achieve our goals and tasks through flexible adaptation. This often requires learning low-complexity structures from the observations, and robustly dealing with outliers of the data. To deal with these challenges, the problem can often be naturally reduced to finding the sparsest vector in a subspace. As an one example, the DPCP approach introduced in \Cref{sec:motivation} has been successfully applied in the context of the three-view problem, which is of fundamental importance in many computer vision applications, such as 3D
reconstruction from 2D images of the scene \cite{tsakiris2018dual}.  

Another successful application of DPCP is on road plane detection from 3D point cloud data using the KITTI
dataset \cite{geiger2013vision}, which is an important computer vision task in autonomous car driving systems. The dataset, recorded from a moving platform while driving in and around Karlsruhe, Germany, consists of image data together with corresponding 3D points collected by a rotating 3D laser scanner. As shown in \Cref{fig:app-dpcp}, one important problem is to determine the 3D points that lie off the road plane (outliers indicated by red) and those on that plane (inliers indicated by blue), follows which the road plane can then be easily estimated. Experimental results in \Cref{fig:app-dpcp} show that the DPCP approach can efficiently estimate the road plane from 3D points including  almost 50\% outliers.

\begin{figure*}[t]
\centering
\begin{minipage}{.24\textwidth}
    \centering
    \includegraphics[width = 0.8\textwidth]{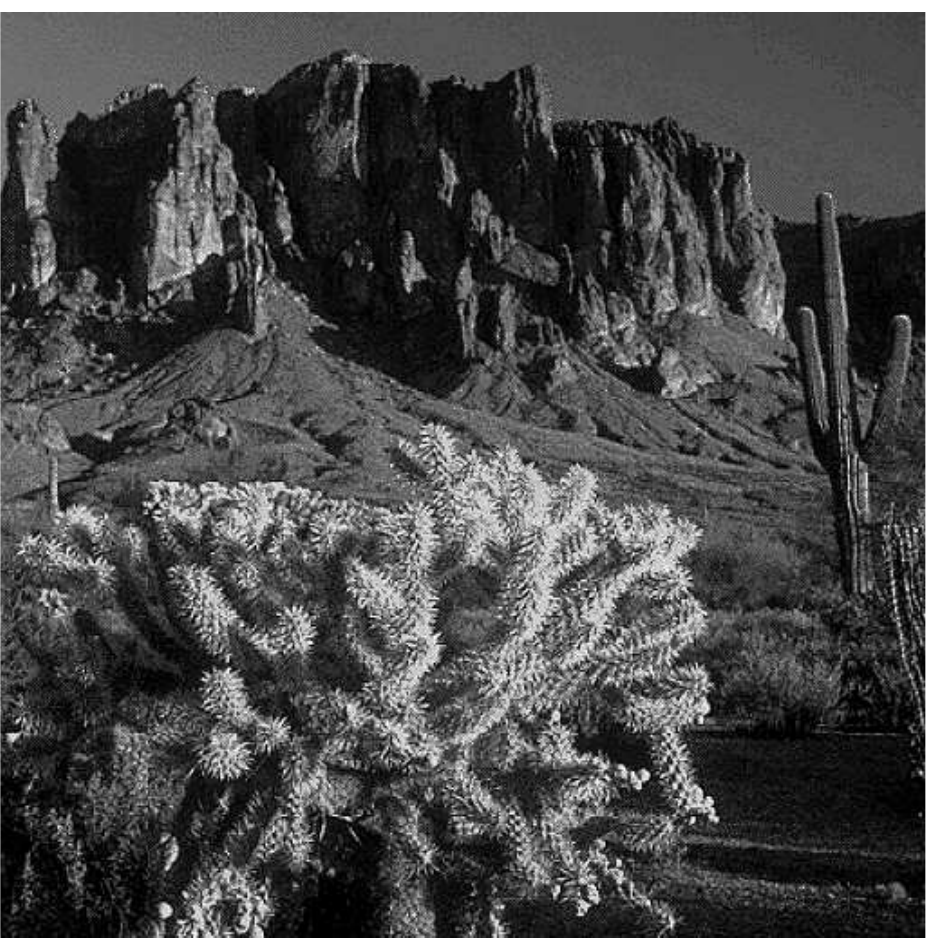}\\
    \vspace{0.05in}
    \includegraphics[width = 0.8\textwidth]{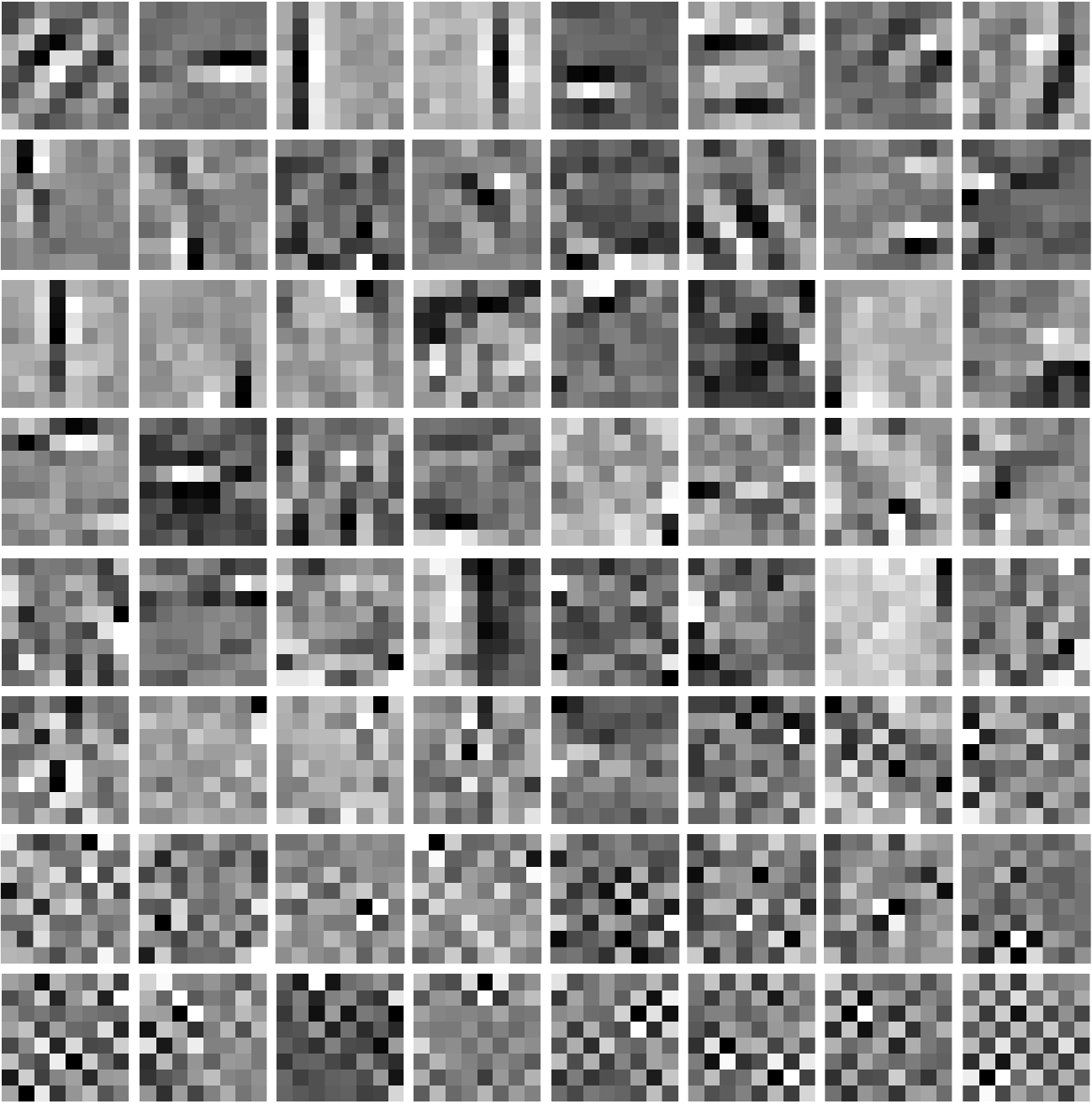} \\
        \vspace{0.05in}
    \includegraphics[width = 0.85\textwidth]{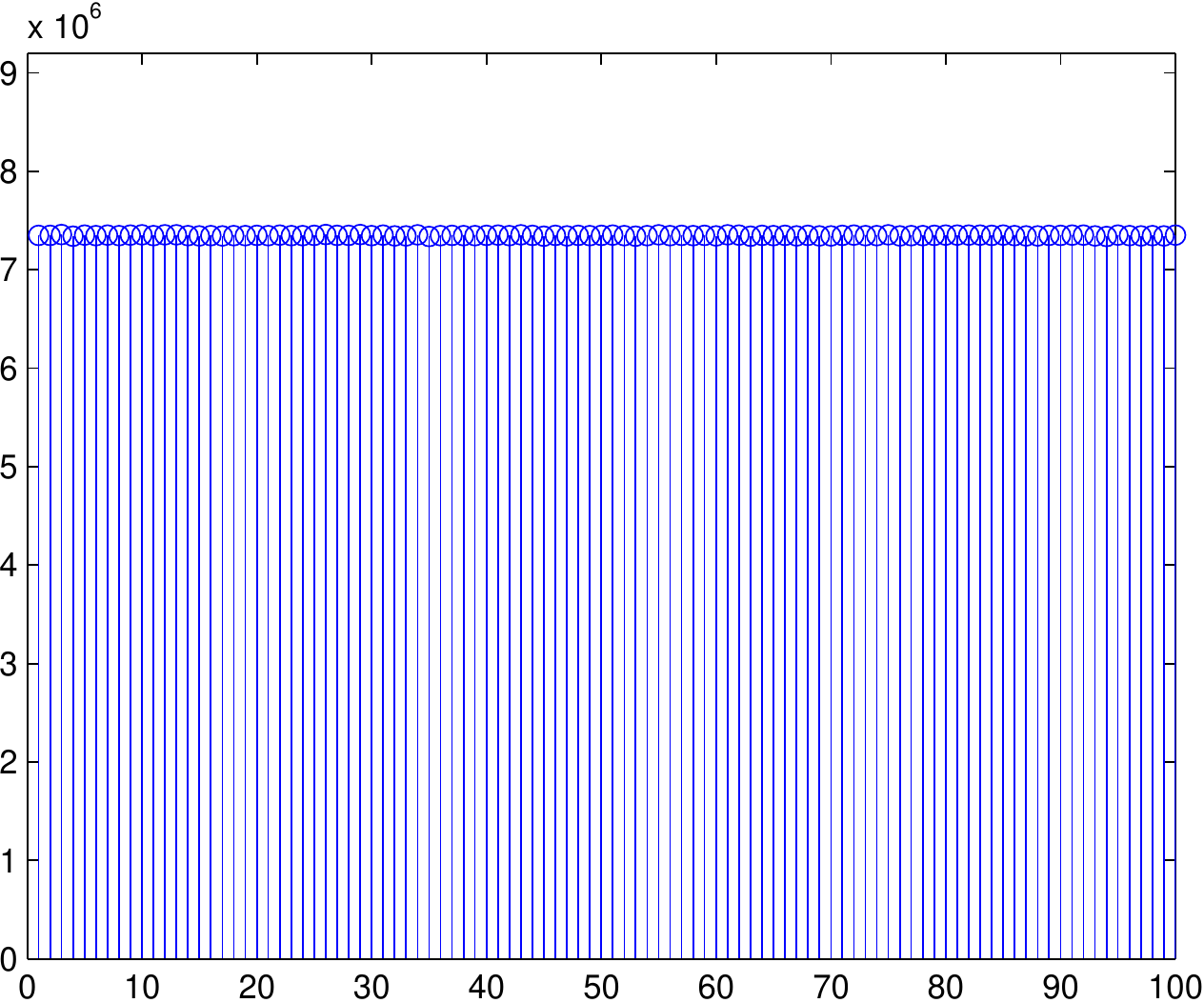}
\end{minipage}%
\begin{minipage}{.24\textwidth}
    \centering
    \includegraphics[width = 0.8\textwidth]{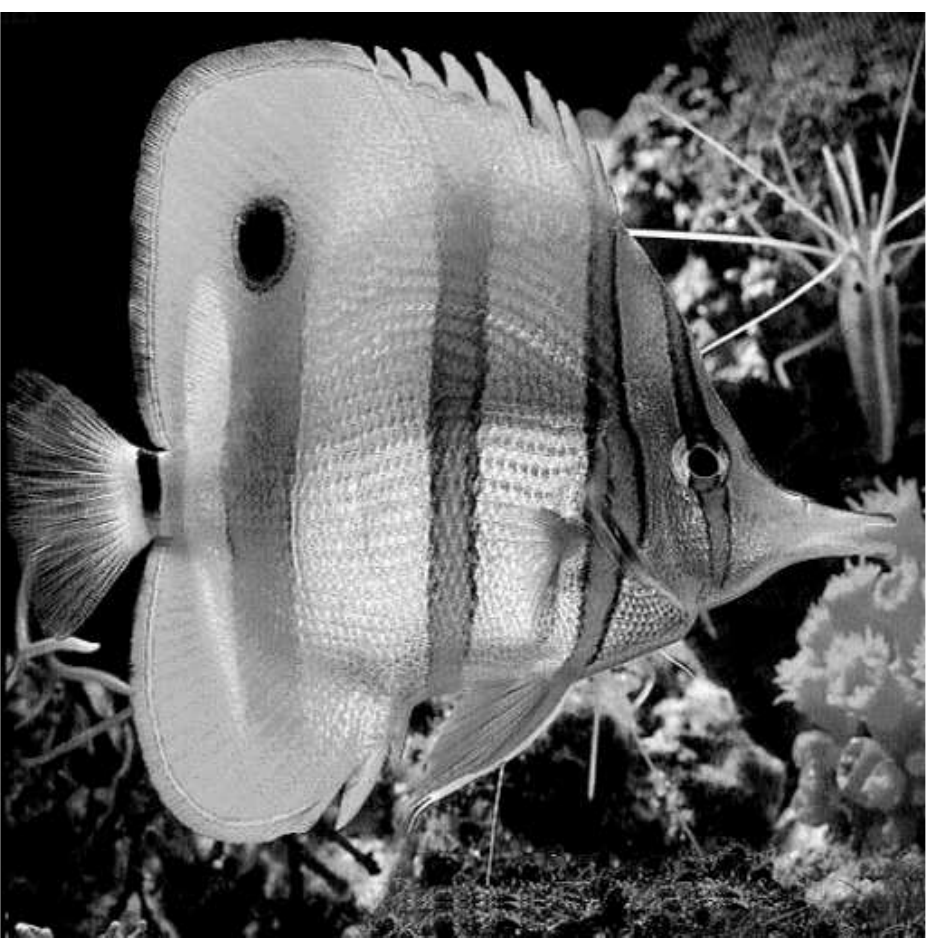}\\
    \vspace{0.05in}
    \includegraphics[width = 0.8\textwidth]{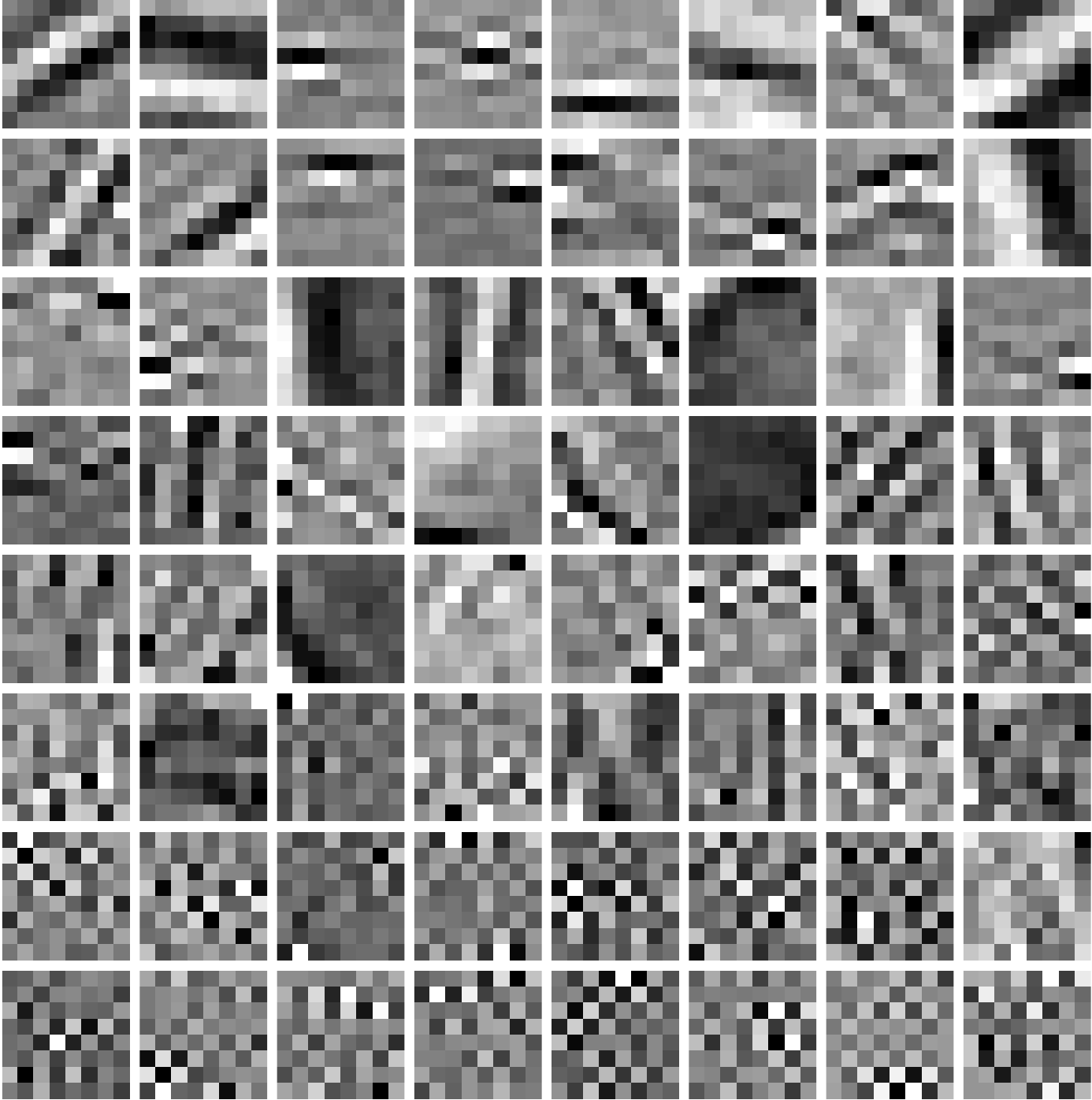} \\
        \vspace{0.05in}
    \includegraphics[width = 0.85\textwidth]{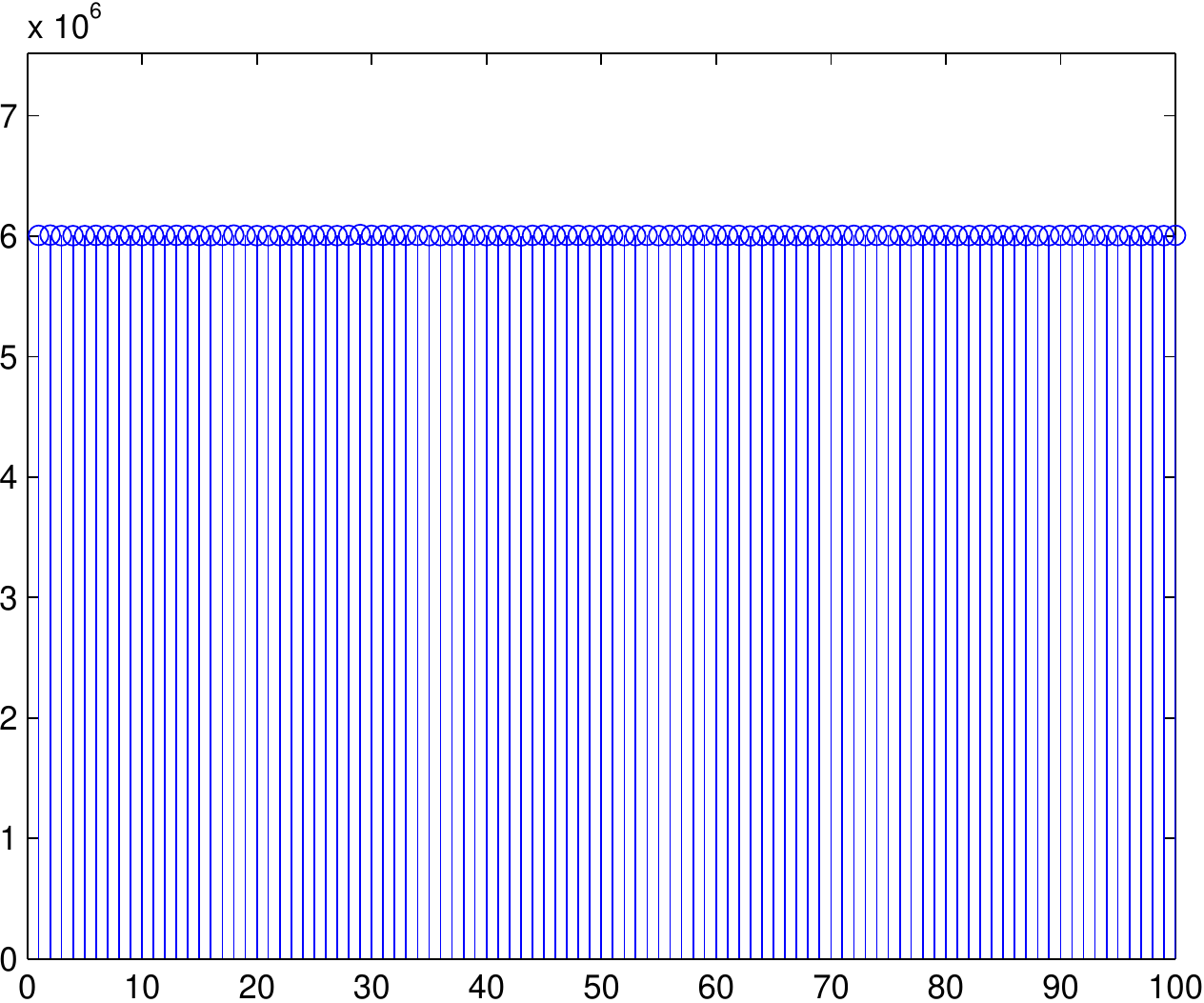}
\end{minipage}%
\begin{minipage}{.24\textwidth}
    \centering
    \includegraphics[width = 0.8\textwidth]{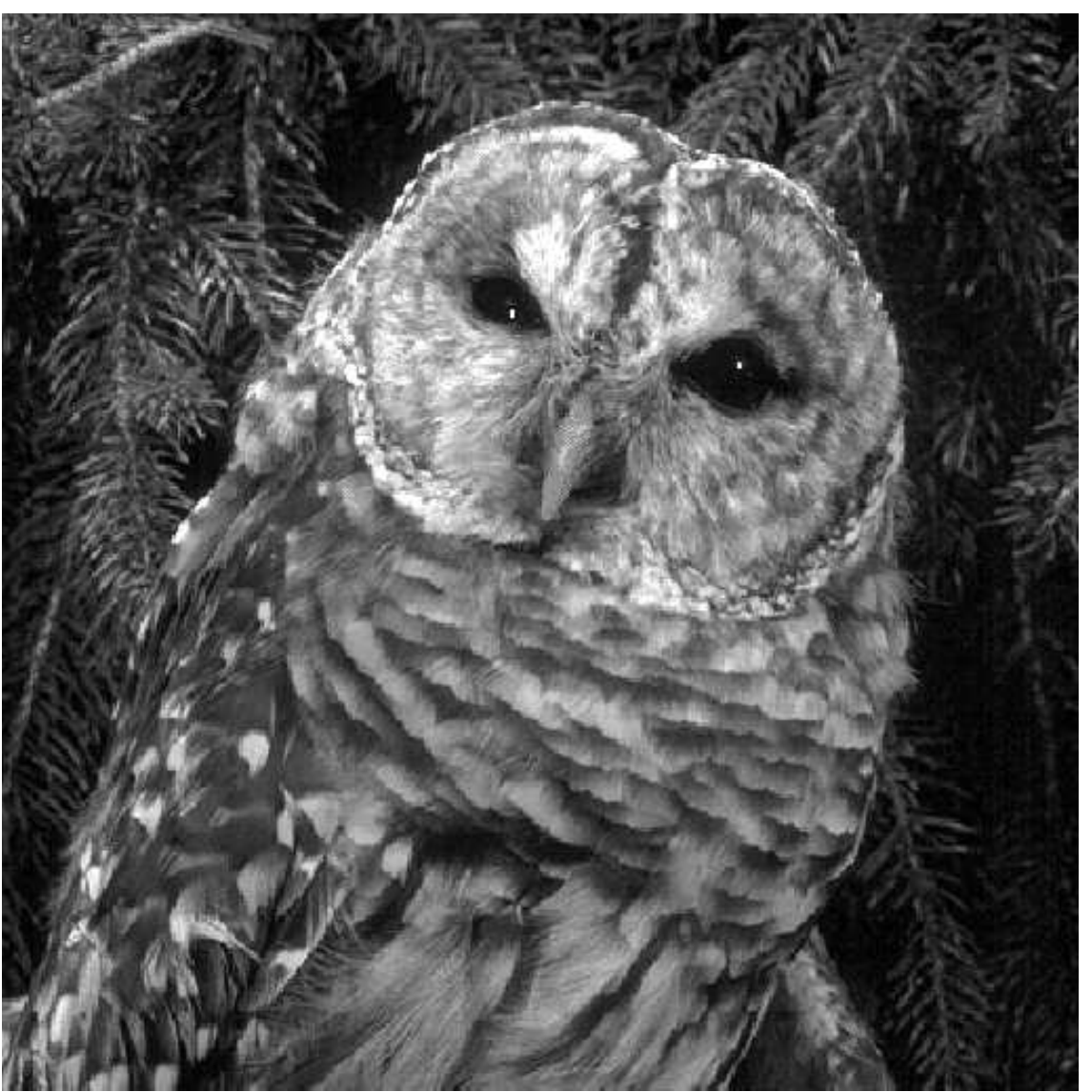}\\
    \vspace{0.05in}
    \includegraphics[width = 0.8\textwidth]{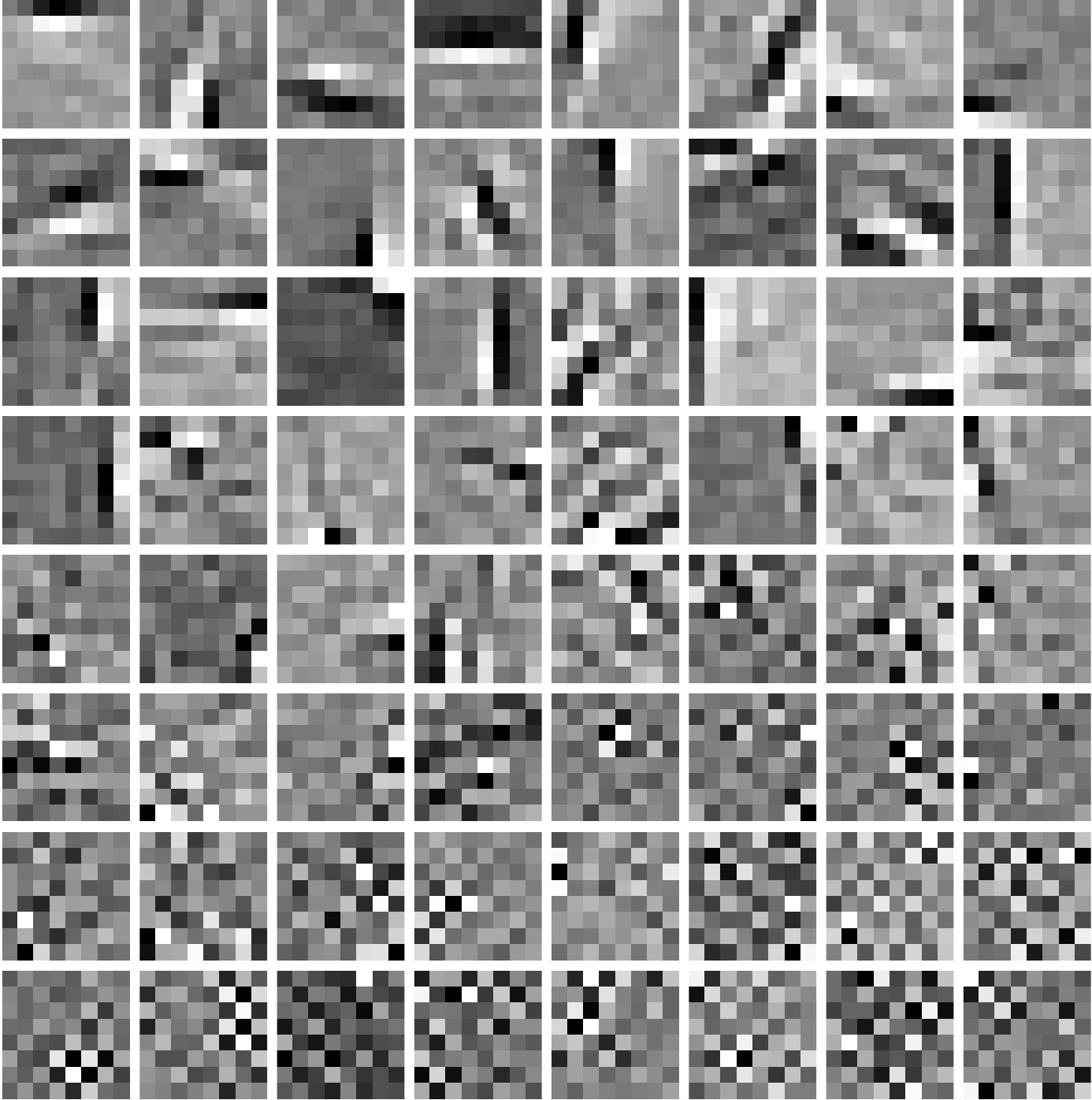}\\
        \vspace{0.05in}
    \includegraphics[width = 0.85\textwidth]{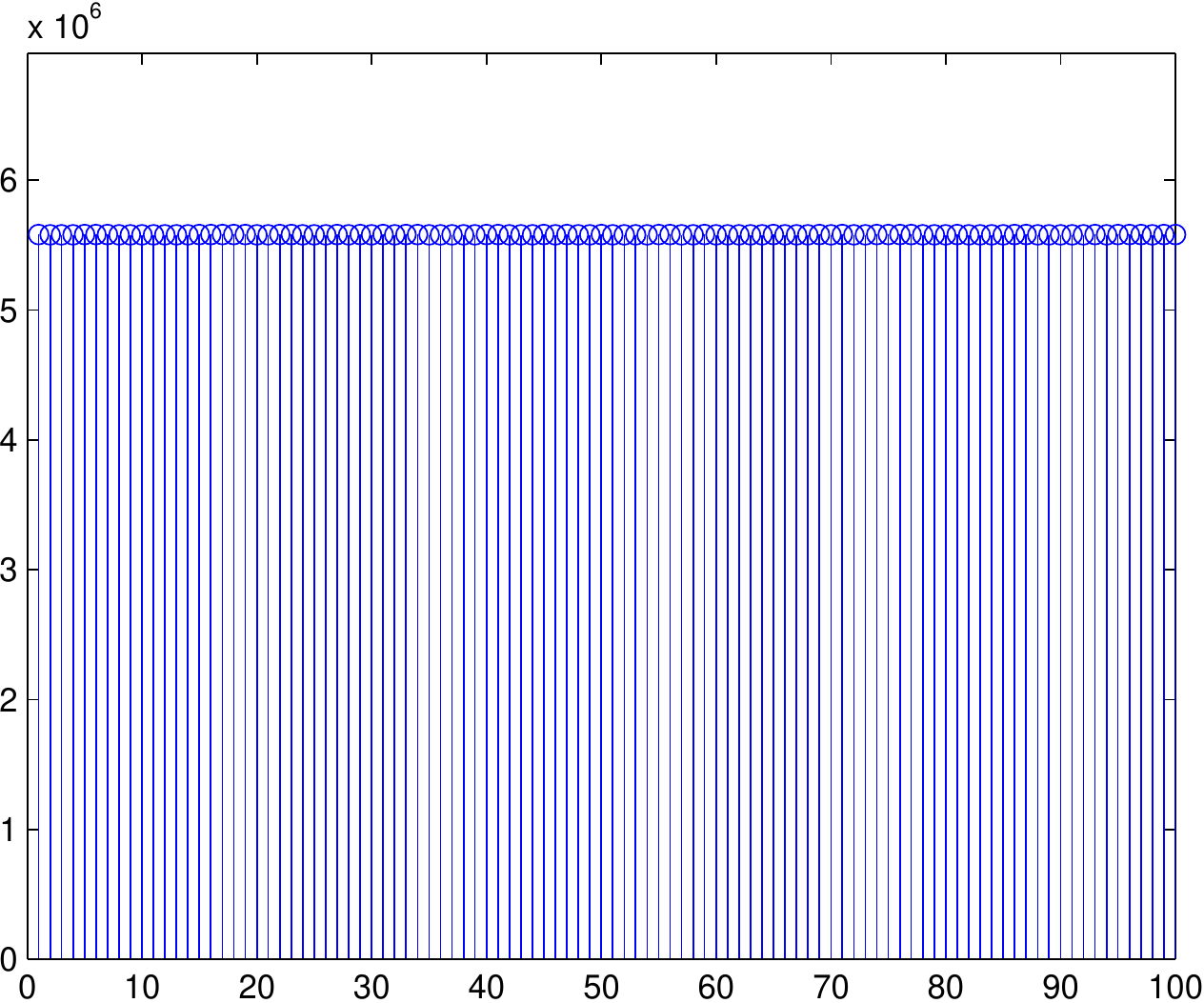}
\end{minipage}
\begin{minipage}{.24\textwidth}
    \centering
    \includegraphics[width = 0.8\textwidth]{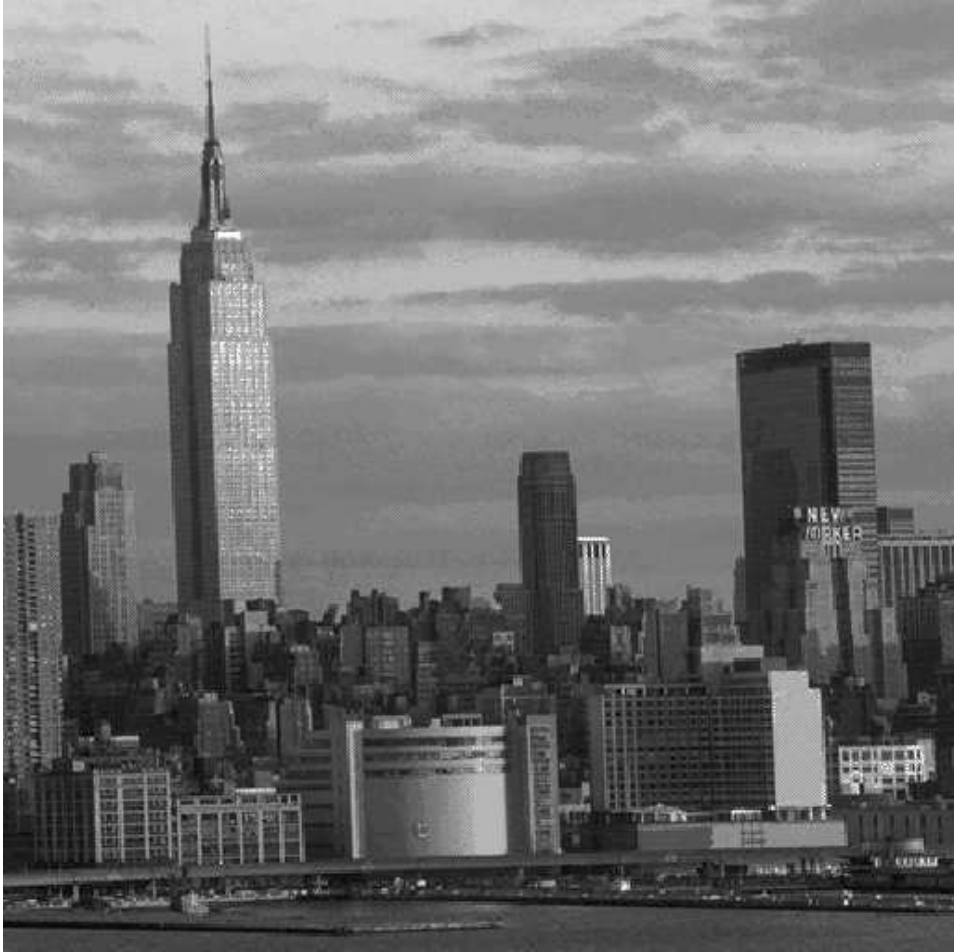}\\
    \vspace{0.05in}
    \includegraphics[width = 0.8\textwidth]{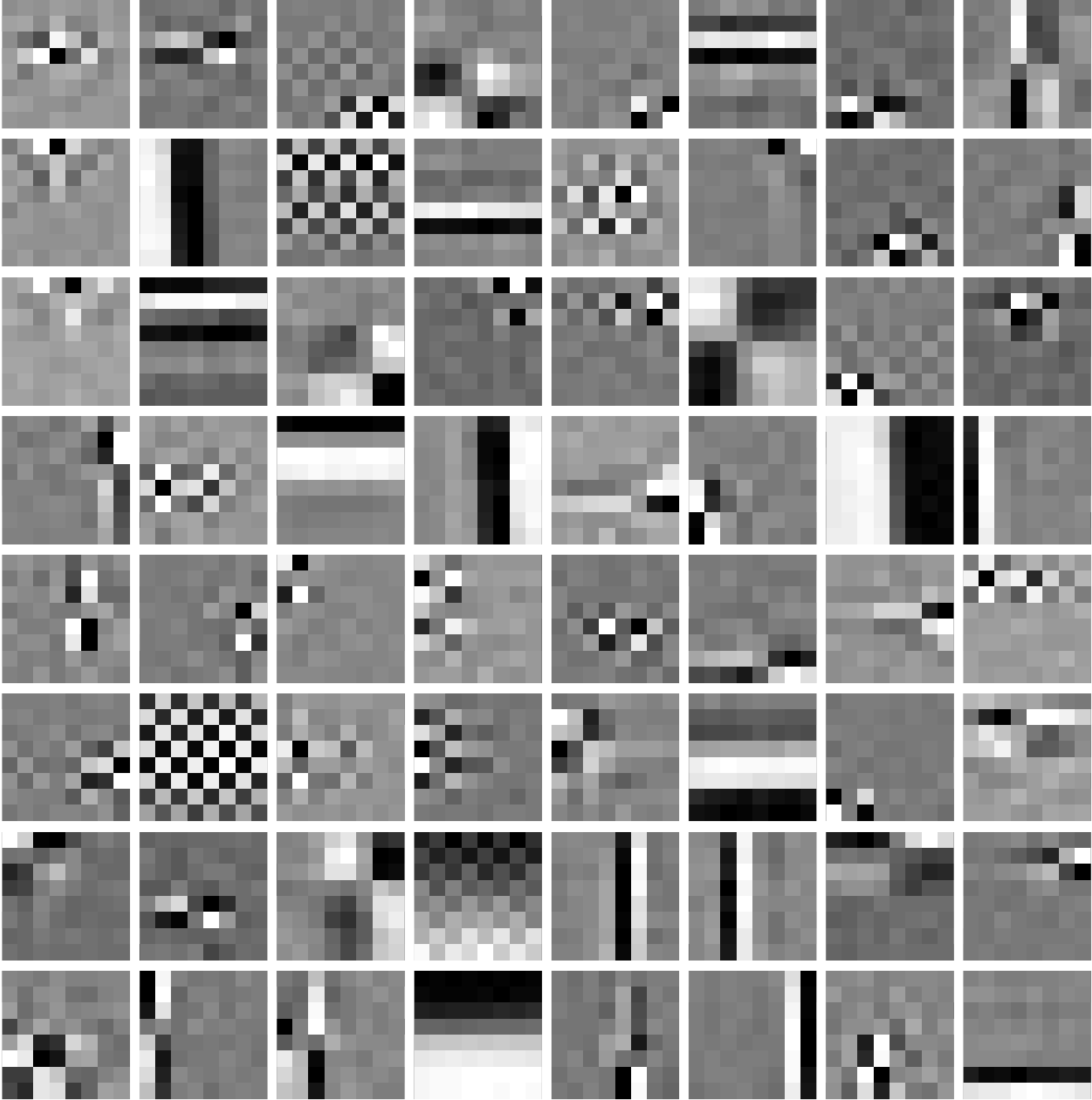} \\
        \vspace{0.05in}
    \includegraphics[width = 0.85\textwidth]{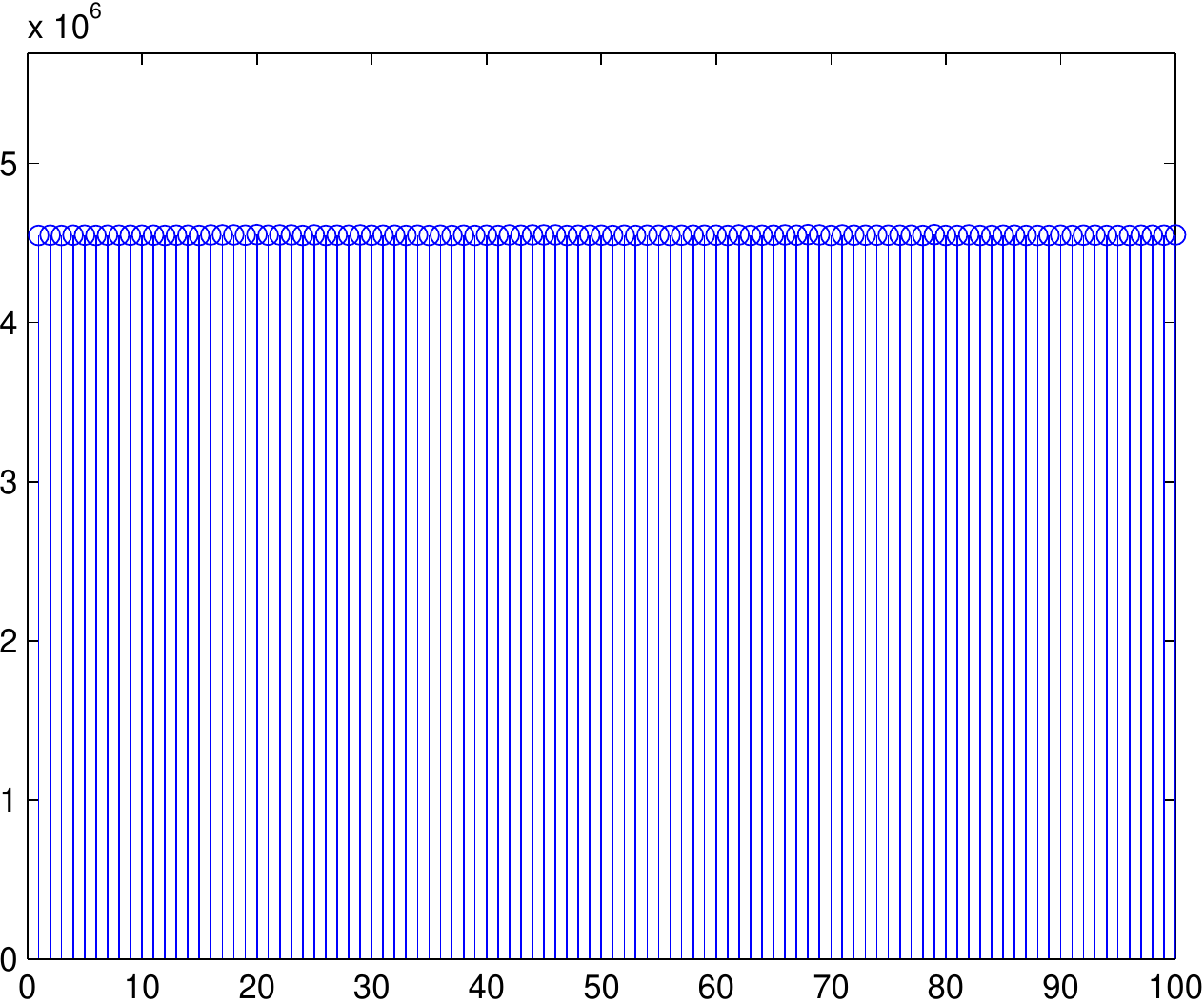}
\end{minipage}
\caption{\textbf{Learning representations of natural images \cite{sun2016complete-ii}.} \textbf{Top:} natural images. \textbf{Middle:} $64$ dictionary elements of size $8\times 8$ learned via solving \eqref{eqn:L1-ncvx} and deflation. \textbf{Bottom:} the plots show the values of $ \norm{\mb A_\star^\top \mb Y}{1} $ across $100$ independent repetitions, where $\mb A_\star$ is the obtained solution for each trial; we observe that regardless of the initialization, the algorithm converges to an equally good point}
\label{fig:app-dl}
\end{figure*}

\paragraph{Representation Learning.}
High dimensional data often contains quite a lot redundant information, and they often possess low-dimensional structures/representations. The performance of modern machine learning and data analytical methods heavily depends on appropriate low-complexity data representations (or features) which capture hidden information underlying the data. While we used to manually craft representations in the past, it has been demonstrated that learned representations from the data show much superior performance \cite{elad2010sparse}. Therefore, (unsupervised) learning of latent representations of high-dimensional data becomes a fundamental problem in signal processing, machine learning, theoretical neuroscience and many other \cite{bengio2013representation}. As alluded in \Cref{sec:motivation}, one of the most important unsupervised representation learning problems is learning sparsely-used dictionaries \cite{olshausen1997sparse}, which aims to learn a compact dictionary such that every data point can be represented by only a few atoms from the dictionary.

However, despite of recent algorithmic and empirical success \cite{wright2010sparse,mairal2014sparse}, most of the methods based on alternating minimizations are lacking theoretical justifications for when and why these algorithms work for dictionary learning. As shown in \Cref{sec:motivation}, when the dictionary is complete, it can be reduced to finding the sparsest vector in a subspace. Moreover, the results in \cite{sun2016complete-i,sun2016complete-ii} showed that this problem can be solved to the target solutions with efficient algorithms and optimal sparsity level. \Cref{fig:app-dl} shows the learned compact representations from natural images using this approach, which is optimized by a second order Riemannian trust region algorithm followed by deflation \cite{sun2016complete-ii}. As we observe, the method does not only enjoy global performance guarantees (see the bottom of \Cref{fig:app-dl}) but also efficiently learn meaningful representation from natural image dataset  (see the middle of \Cref{fig:app-dl}).

\paragraph{Scientific Imaging.} In many imaging science applications, we often face the problem of recovering a low-complexity signal provided observations taken from an unknown physical system. For instance, in fluorescent optical microscopy imaging, super-resolution microscopy is a new computation based imaging technique which breaks the resolution limits of conventional optical fluorescence microscopy \cite{betzig2006imaging,hess2006ultra,rust2006sub}. The basic principle is using photoswitchable florescent probes to create multiple sparse frames of individual molecules to temporally separate the spatially overlapping low resolution image. To improve the resolution limit, we need to computationally recover a sequence of sparse high resolution (HR) images from their convolution with a point spread function (i.e., low resolution images). However, in many scenarios (especially in 3D imaging), as it is often difficult to directly estimate the PSF due to defocus and unknown aberrations \cite{sarder2006deconvolution}, it is more desired to jointly estimate both the PSF and high resolution images by solving a sparse blind deconvolution problem with multiple inputs.

\begin{figure*}[t]
	\centering
	\centering
    \begin{minipage}[c]{0.24\textwidth}
    	\centering
    	\includegraphics[width = 0.9\linewidth]{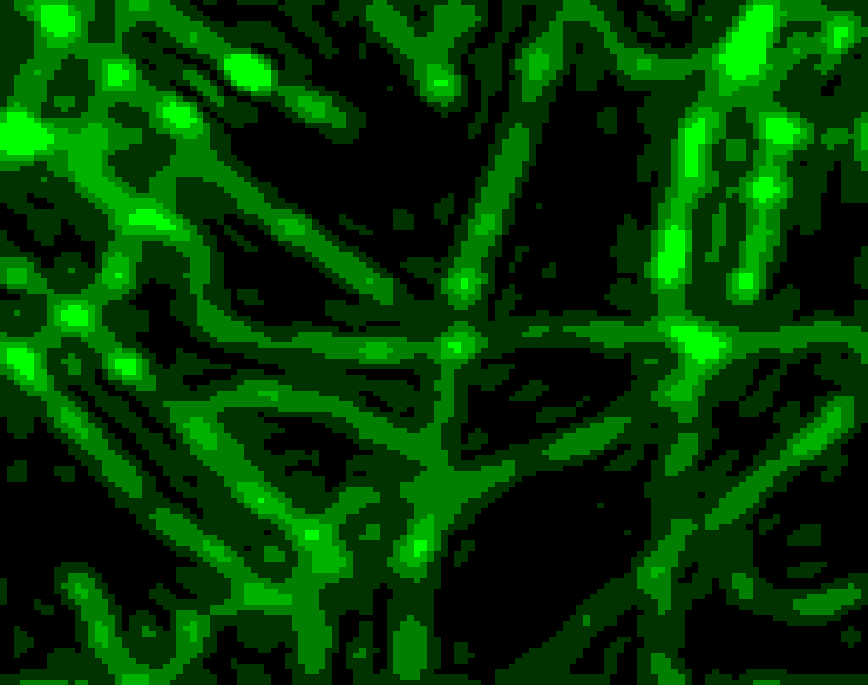}
    	\subcaption{Observation}
    \end{minipage}
    \begin{minipage}[c]{0.24\textwidth}
    	\centering
    	\includegraphics[width = 0.9\linewidth]{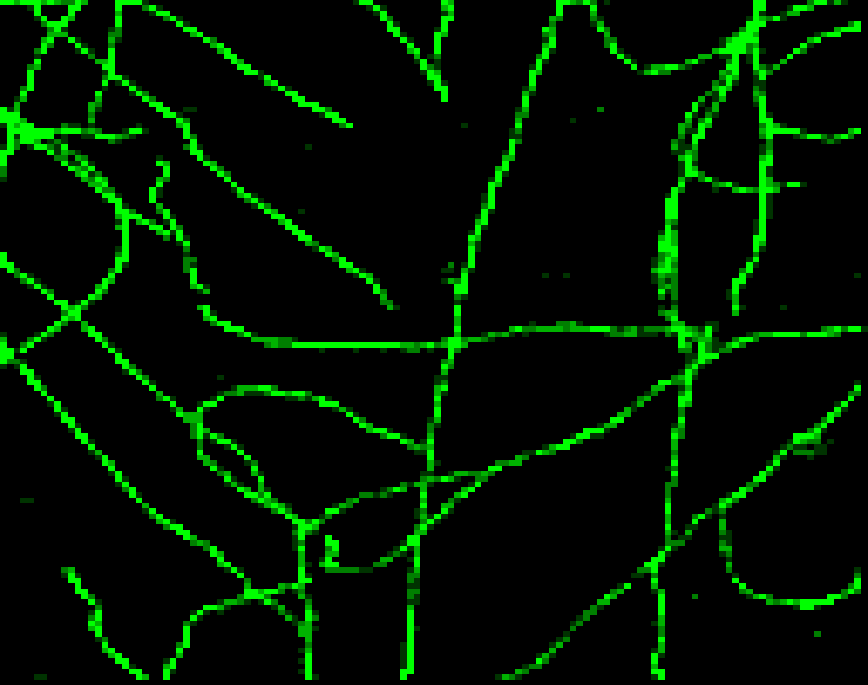}
    	\subcaption{HR Image: Truth}
    \end{minipage}
    \begin{minipage}[c]{0.24\textwidth}
    	\centering
    	\includegraphics[width = 0.9\linewidth]{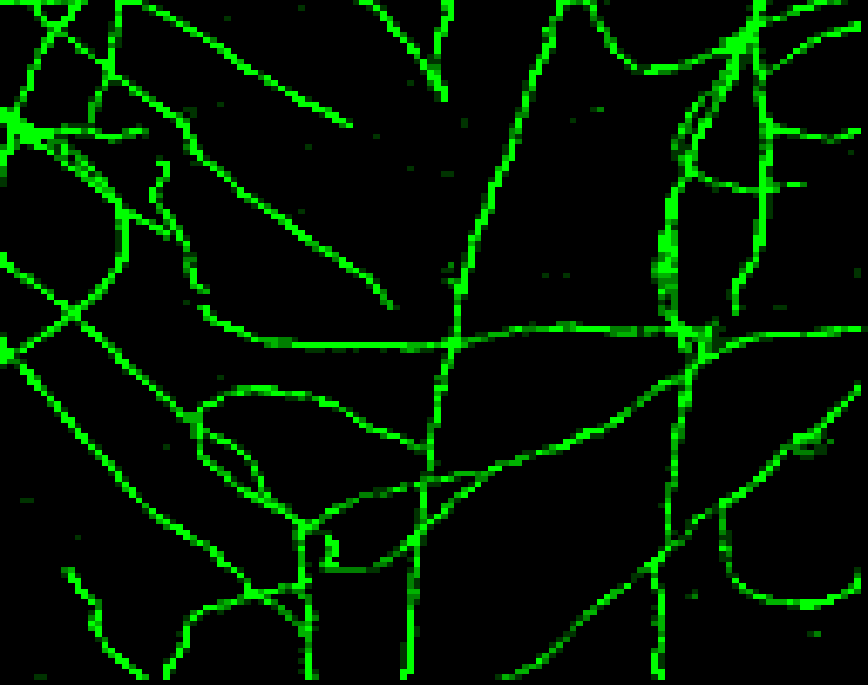}
    	\subcaption{HR Image: Recovered}
    \end{minipage} \\ 
	\begin{minipage}[c]{0.32\textwidth}
		\centering
		\includegraphics[width = 0.7\linewidth]{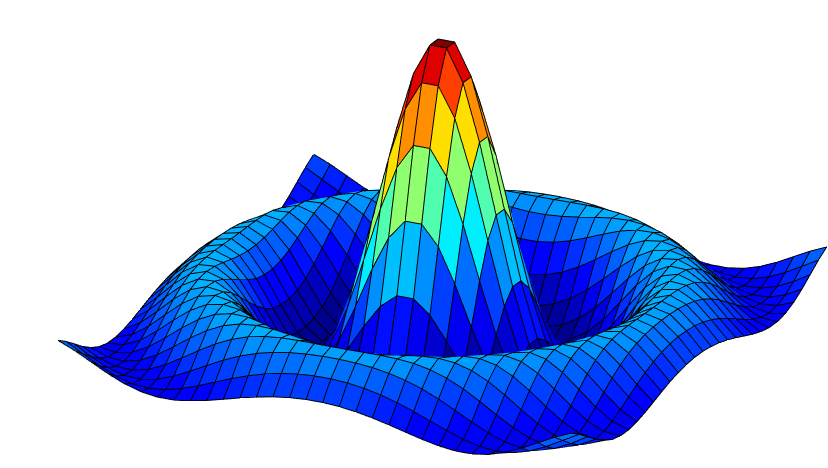}
		\subcaption{PSF: Ground truth}
	\end{minipage}
	\begin{minipage}[c]{0.32\textwidth}
		\centering
		\includegraphics[width = 0.7\linewidth]{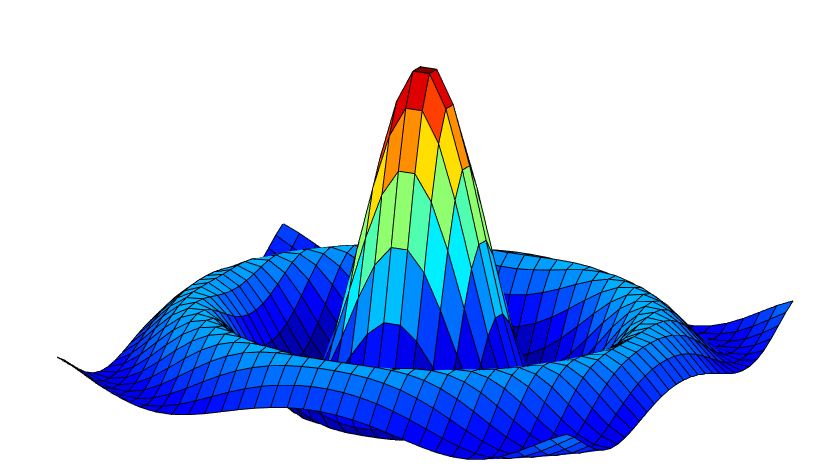}
		\subcaption{PSF: Recovered}
	\end{minipage}
	\caption{\textbf{Solving sparse blind deconvolution for solving super-resolution microscopy imaging \cite{qu2019blind}.} Results on a standard stochastic optical reconstruction microscopy \cite{rust2006sub} dataset. \textbf{Top:} from left to right, observed blurred image, ground truth and recovered HR images \textbf{Bottom:} from left to right, ground truth and recovered PSFs. }
	\label{fig:app-bd}
\end{figure*}

As discussed in \Cref{sec:motivation}, this sparse blind deconvolution problem can be reduced to finding the sparsest vector in a subspace, which can be efficiently solved by the algorithms in \Cref{sec:optimization}. As a demonstration of effectiveness, we test this approach\footnote{Here, we consider the Huber-loss for $\varphi$, and solve the problem via RGD.} on a realistic simulated dataset obtained from SMLM challenge website\footnote{Available at \url{http://bigwww.epfl.ch/smlm/datasets/index.html?p=tubulin-conjal647}.} using $1000$ video frames. The fluorescence wavelength is 690 nanometer (nm) and the imaging frequency is $f=25Hz$. Each frame is of size $128\times 128$ with 100 nm pixel resolution, and we solve the single-molecule localization problem on the same grid\footnote{Here, we are estimating the HR images on the same grid as the original image. To obtain even higher resolution than the result we obtain here, people are usually estimating the HR images on a finer grid.}. As observed in \Cref{fig:app-bd}, by reducing and solving the finding the sparsest vector in a subspace problem using simple algorithms, it can near perfectly recover both the underlying PSF and HR images, producing accurate recovery results. 

\section{Conclusion and Future Directions}\label{sec:conclusion}

This work is part of a recent surge of research efforts on deriving provable and practical nonconvex algorithms to central problems in modern signal processing and machine learning. In this paper, we reviewed several important aspects of recent advances on nonconvex optimization methods for solving the problem of finding the sparsest vector in a subspace, ranging from problem formulation, geometric analysis of optimization landscapes, to efficient algorithms and applications. In the following, we discuss several open problems to be addressed along this line of research in the near future.

\paragraph{Towards more disciplined nonconvex optimization theory.} Despite of recent theoretical and algorithmic advances, our understandings of nonconvex optimization is still \emph{far} from satisfactory --- the current analysis is delicate, case-by-case, and pertains to problems with elementary symmetry (e.g., permutation or shift symmetry) and simple manifold (e.g., sphere). Analogous to the study of convex functions \cite{boyd2004convex}, there is a pressing need for simpler analytic tools, to identify and generalize benign properties for new nonconvex problems appearing in signal processing and machine learning.

\paragraph{Learning low-complexity structures over more complicated manifold.} In this work, we formulate the problems such as robust subspace recovery and dictionary learning as finding a sparse vector in a subspace, which is constrained over the sphere. However, more natural and robust formulations for these problems involves optimization over more complicated manifolds, such as Stiefel manifold. More technical tools need to be developed towards a better understanding of optimization over these complicated manifolds, despite recent endeavors \cite{zhu2019linearly,li2019nonsmooth,hu2019brief,zhai2019complete}.

\paragraph{Applications.} In this paper, we reviewed a variety of optimization algorithm for finding a sparse vector in a subspace, with global theoretical guarantees. Moreover, these algorithms are practical for handling large dataset as we demonstrated on several applications including machine intelligence, representation learning, and imaging sciences. However, we believe the potential of seeking sparse/structured element in a subspace is still largely unexplored, despite the cases we mentioned and demonstrated in this work. We hope the motivating applications discussed in this survey could inspire more application ideas of these results.

\section*{Acknowledgement}
 QQ thanks the generous support of the Microsoft graduate research fellowship and Moore-Sloan fellowship. ZZ and RV are partly supported by NSF IIS 1704458.
 XL would like to acknowledge the support by Grant CUHK14210617 from the Hong Kong Research Grants Council. JW acknowledges the support by NSF CCF 1527809 and NSF IIS 1546411.

\section*{Author Biography}

\noindent \textbf{Qing Qu} (qq213@nyu.edu) is a Moore-Sloan data science fellow at the Center for Data Science, New York University. He received his Ph.D from Columbia University in Electrical Engineering in Oct. 2018. He received his B.Eng. from Tsinghua University in Jul. 2011, and a M.Sc.from the Johns Hopkins University in Dec. 2012, both in Electrical and Computer Engineering. He interned at U.S. Army Research Laboratory in 2012 and Microsoft Research in 2016, respectively. His research interest lies at the intersection of signal/image processing, machine learning, numerical optimization, with focus on developing efficient nonconvex methods and global optimality guarantees for solving engineering problems in signal processing, computational imaging, and machine learning. He is the recipient of Best Student Paper Award at SPARS’15 (with Ju Sun, John Wright), and the recipient of 2016-18 Microsoft Research Fellowship in North America.
\par\smallskip
\noindent \textbf{Zhihui Zhu} (zhihui.zhu@du.edu) received the B.Eng. degree in communication engineering from the Zhejiang University of Technology, Hangzhou, China, in 2012, and the Ph.D. degree in electrical engineering from the Colorado School of Mines, Golden, CO, USA, in 2017. He was a Postdoctoral Fellow in the Mathematical Institute for Data Science at the Johns Hopkins University, Baltimore, MD, in 2018-2019. He is currently an Assistant Professor with the Department of Electrical and Computer Engineering, University of Denver, CO. His research interests include exploiting inherent structures and applying optimization methods with guaranteed performance for signal processing, machine learning, and data analysis.
\par\smallskip
\noindent \textbf{Xiao Li} (xli@ee.cuhk.edu.hk) received B.Eng. degree in communication engineering with the Zhejiang University of Technology in 2016. He is currently pursuing his Ph.D. degree at The Chinese University of Hong Kong.  His current research interests lie in utilizing  nonconvex formulation and provable algorithms to problems arising from signal processing, machine learning and data science in order for fast and efficient computation.
\par\smallskip
\noindent \textbf{Manolis C. Tsakiris} (mtsakiris@shanghaitech.edu.cn) is an electrical engineering and computer science graduate of the National Technical University of Athens, Greece. He holds an MS degree in signal processing from Imperial College London, UK, and a PhD degree in theoretical machine learning from Johns Hopkins University, USA, under the supervision of Prof. Ren\'e Vidal. Since August 2017 he is an assistant professor at the School of Information Science and Technology (SIST) at ShanghaiTech University. He pursues research on fundamental aspects of subspace learning and related problems in commutative algebra.
\par\smallskip
\noindent \textbf{John N. Wright} (jw2966@columbia.edu) received his B.S. degree in computer engineering, his M.S. degree in electrical engineering, and his Ph.D. degree in electrical engineering from the University of Illinois at Urbana–Champaign (UIUC) in 2004, 2007, and 2009, respectively. From 2009 to 2011, he was with Microsoft Research Asia. He is currently an associate professor in the Electrical Engineering Department at Columbia University, New York. His has received a number of awards, including the 2012 Conference on Learning Theory Best Paper Award (with Dan Spielman and Huan Wang), the 2009 Lemelson-Illinois Prize for Innovation for his work on face recognition, and the 2009 UIUC Martin Award for Excellence in Graduate Research. His research interests include high-dimensional data analysis. He is a Member of the IEEE. 
\par\smallskip	
\noindent \textbf{Ren\'e Vidal} (rvidal@jhu.edu) received the B.S. degree in electrical engineering (highest honors) from the Pontificia Universidad Catolica de Chile, Santiago, Chile, in 1997 and the M.S. and Ph.D. degrees in electrical engineering and computer sciences from the University of California, Berkeley, CA, USA, in 2000 and 2003, respectively. He was a Research Fellow at the National ICT Australia in the fall of 2003 and has been a faculty member in the Department of Biomedical Engineering and the Center for Imaging Science of The Johns Hopkins University since 2004. He is co-author of the book “Generalized Principal Component Analysis” (2016), co-editor of the book “Dynamical Vision,” and co-author of more than 200 articles in machine learning, computer vision, biomedical image analysis, hybrid systems, robotics, and signal processing. He is or has been Associate Editor of Medical Image Analysis, the IEEE Transactions on Pattern Analysis and Machine Intelligence, the SIAM Journal on Imaging Sciences, Computer Vision and Image Understanding, and the Journal of Mathematical Imaging and Vision, and a Guest Editor of the International Journal on Computer Vision and Signal Processing Magazine. He is a member of the ACM and SIAM.

{\small
\medskip
\bibliographystyle{ieeetr}
\bibliography{finding-ncvx,finding-TIT}
}

\end{document}